\documentclass[running heads]{llncs}
\usepackage{cite}
\usepackage{amsmath}
\usepackage{algorithmic}
\usepackage{array}
\usepackage{mdwmath}
\usepackage{mdwtab}
\usepackage{eqparbox}
\usepackage{epsfig}
\usepackage[most]{tcolorbox}
\usepackage[caption=false]{subfig}
\usepackage{float}
\usepackage{url}
\usepackage{soul}
\usepackage{enumerate}
\usepackage{amssymb}
\usepackage[breaklinks]{hyperref}
\usepackage{hyperref}
\hypersetup{colorlinks=true, urlcolor=blue}

\begin{document}

\title{IIIT-AR-13K: A New Dataset for Graphical Object Detection in Documents}

\author{Ajoy Mondal\inst{1} \and
Peter Lipps\inst{2} \and
C V Jawahar\inst{1}}

\institute{Centre for Visual Information Technology, \\
International Institute of Information Technology, Hyderabad, India \\
\email{\{ajoy.mondal,jawahar\}@iiit.ac.in}
\and
Open Text Software GmbH, Grasbrunn/Munich, Germany\\
\email{peter.lipps@opentext.com}}

\maketitle              

\begin{abstract}
We introduce a new dataset for graphical object detection in business documents, more specifically annual reports. This dataset, {\sc iiit-ar}-13{\sc k}, is created by manually annotating the bounding boxes of graphical or page objects in publicly available annual reports. This dataset contains a total of 13{\sc k} annotated page images with objects in five different popular categories --- table, figure, natural image, logo, and signature. It is the largest manually annotated dataset for graphical object detection. Annual reports created in multiple languages for several years from various companies bring high diversity into this dataset. We benchmark {\sc iiit-ar}-13{\sc k} dataset with two state of the art graphical object detection techniques using {\sc f}aster {\sc r-cnn}~\cite{ren2015faster} and {\sc m}ask {\sc r-cnn}~\cite{he2017mask} and establish high baselines for further research. Our dataset is highly effective as training data for developing practical solutions for graphical object detection in both business documents and technical articles. By training with {\sc iiit-ar}-13{\sc k}, we demonstrate the feasibility of a single solution that can report superior performance compared to the equivalent ones trained with a much larger amount of data, for table detection. We hope that our dataset helps in advancing the research for detecting various types of graphical objects in business documents\footnotemark[1].  
\footnotetext[1]{\url{http://cvit.iiit.ac.in/usodi/iiitar13k.php}}

\end{abstract}

\section{Introduction}

Graphical objects such as tables, figures, logos, equations, natural images, signatures, play an important role in the understanding of document images. They are also important for information retrieval from document images. Each of these graphical objects contains valuable information about the document in a compact form. Localizing such graphical objects is a primary step for understanding documents or information extraction/retrieval from the documents. Therefore, the detection of those graphical objects from the document images has attracted a lot of attention in the research community~\cite{gilani2017table,schreiber2017deepdesrt,tran2015table,hao2016table,kavasidis2019saliency,tablebank,Siddiqui2018DeCNTDD,melinda_2019,li_2019,yilun_2019,sun_2019,ranajit_2019}. Large diversity within each category of the graphical objects makes detection task challenging. Researchers have explored a variety of algorithms for detecting graphical objects in the documents. Numerous benchmark datasets are also available in this domain to evaluate the performance of newly developed algorithms. In this paper, we introduce a new dataset, {\sc iiit-ar-13k} for graphical object detection.

Before we describe the details of the new dataset, we make the following observations:
\begin{enumerate}
\item State of the art algorithms (such as~\cite{gilani2017table,schreiber2017deepdesrt,Siddiqui2018DeCNTDD,sun_2019,yilun_2019,kavasidis2019saliency,ranajit_2019}) for graphical object detection are motivated by the success of object detection in computer vision (such as {\sc f}aster {\sc r-cnn} and {\sc m}ask {\sc r-cnn}). However, the high accuracy expectations in documents (similar to the high accuracy expectations in {\sc ocr}) make the graphical object detection problem, demanding and thereby different compared to that of detecting objects in natural images.
\item With documents getting digitized extensively in many business workflows, automatic processing of business documents has received significant attention in recent years. However, most of the existing datasets for graphical object detection are still created from scientific documents such as technical papers.
\item In recent years, we have started to see large automatically annotated or synthetically created datasets for graphical object detection. We hypothesize that, in high accuracy regime, carefully curated small data adds more value than automatically generated large datasets. At least, in our limited setting, we validate this hypothesis later in this paper.
\end{enumerate}

Popular datasets for graphical object detection (or more specifically table detection) are {\sc icdar 2013} table competition dataset~\cite{gobel2013icdar} (i.e., {\sc icdar-2013}), {\sc icdar 2017} competition on page object detection dataset~\cite{icdar_2017} (i.e., {\sc icdar-pod-2017}), c{\sc td}a{\sc r}~\cite{icdar_table_2019}, 
{\sc unlv}~\cite{unlv}, {\sc m}armot table recognition dataset~\cite{marmot}, {\sc d}eep{\sc f}igures~\cite{siegel2018extracting}, {\sc p}ub{\sc l}ay{\sc n}et~\cite{publaynet}, and {\sc t}able{\sc b}ank~\cite{tablebank}.
Most of these existing datasets are limited with respect to their size (except {\sc d}eep{\sc f}igures, {\sc p}ub{\sc l}ay{\sc n}et, and {\sc t}able{\sc b}ank) or category labels (except {\sc icdar-pod}-2017, {\sc d}eep{\sc f}igures, and {\sc p}ub{\sc l}ay{\sc n}et). Most of them (except {\sc icdar-2013}, c{\sc td}a{\sc r}, and {\sc unlv}) consist of only scientific research articles, handwritten documents, and e-books. Therefore, they are limited in variations in layouts and structural variations in appearances. 

On the contrary, business documents such as annual reports, pay slips, bills, receipt copies, etc. of the various companies are more heterogeneous to their layouts and complex visual appearance of the graphical objects. In such kind of heterogeneous documents, the detection of graphical objects becomes further difficult. 

In this paper, we introduce a new dataset, named {\sc iiit-ar}-13{\sc k} for localizing graphical objects in the annual reports (a specific type of business documents) of various companies. For this purpose, we randomly select publicly available annual reports in English and other languages (e.g., French, Japanese, Russian, etc.) of multiple (more than ten) years of twenty-nine different companies. We manually annotate the bounding boxes of five different categories of graphical objects (i.e., table, figure, natural image, logo, and signature), which frequently appear in the annual reports. {\sc iiit-ar}-13{\sc k} dataset contains 13{\sc k} annotated pages with 16{\sc k} tables, 3{\sc k} figures, 3{\sc k} natural images, 0.5{\sc k} logos, and 0.6{\sc k} signatures. To the best of the author's knowledge, the newly created dataset is the largest among all the existing datasets where ground truths are annotated manually for detecting graphical objects (more than one category) in documents. 

We use {\sc f}aster {\sc r-cnn}~\cite{ren2015faster} and {\sc m}ask {\sc r-cnn}~\cite{he2017mask} algorithms to benchmark the newly created dataset for graphical object detection task in business documents. Experimentally, we observe that the creation of a model trained with {\sc iiit-ar}-13{\sc k} dataset achieves better performance than the model trained with the larger existing datasets. From the experiments, we also observe that the model trained with the larger existing datasets achieves the best performance by fine-tuning with a small number (only 1{\sc k}) of images from {\sc iiit-ar}-13{\sc k} dataset.   

Our major contributions/claims are summarized as follows:

\begin{itemize}
\item We introduce a highly variate new dataset for localizing graphical objects in documents. The newly created dataset is the largest among the existing datasets where ground truth is manually annotated for the graphical object detection task. It also has a larger label space compared to most existing datasets. 

\item We establish {\sc f}aster {\sc r-cnn} and {\sc m}ask {\sc r-cnn} based benchmarks on the popular datasets. We report very high quantitative detection rates on the popular benchmarks.

\item Though smaller than some of the recent datasets, this dataset is more effective as training data for the detection task due to the inherent variations in the object categories. This is empirically established by creating a unique model trained with {\sc iiit-ar}-13{\sc k} dataset to detect graphical objects in all existing datasets. 

\item Models trained with the larger existing datasets achieve the best performance after fine-tuning with a very limited number (only 1{\sc k}) of images from {\sc iiit-ar}-13{\sc k} dataset.
\end{itemize}

\section{Preliminaries}

Graphical objects such as tables, various types of figures (e.g., bar chart, pie chart, line plot, natural image, etc.) equations, and logos in documents contain valuable information in a compact form. Understanding of document images 
requires localizing of such graphical objects as an initial step. Several datasets (e.g., {\sc icdar-2013}~\cite{gobel2013icdar}, {\sc icdar-pod-2017}~\cite{icdar_2017}, c{\sc td}a{\sc r}~\cite{icdar_table_2019}, {\sc unlv}~\cite{unlv}, {\sc m}armot~\cite{marmot}, {\sc d}eep{\sc f}igures~\cite{siegel2018extracting}, {\sc p}ub{\sc l}ay{\sc n}et~\cite{publaynet}, and {\sc t}able{\sc b}ank~\cite{tablebank}) exist in the literature, which are dedicated to localize graphical objects (more specifically tables) in the document images. Ground truth bounding boxes of {\sc icdar-2013}, {\sc icdar-pod-2017}, c{\sc td}a{\sc r}, {\sc unlv}, and {\sc m}armot are annotated manually. Ground truth bounding boxes of {\sc d}eep{\sc f}igures, {\sc p}ub{\sc l}ay{\sc n}et, and {\sc t}able{\sc b}ank are generated automatically. Among these datasets, only {\sc icdar-pod-2017}, {\sc d}eep{\sc f}igures, and {\sc p}ub{\sc l}ay{\sc n}et are aimed to address localizing a wider class of graphical objects. Other remaining datasets are designed only for localizing tables. Some of these datasets (e.g., {\sc icdar-2013}, c{\sc td}a{\sc r}, {\sc unlv}, and {\sc t}able{\sc b}ank) are also used for table structure recognition and table recognition tasks in the literature. Some other existing datasets such as {\sc s}ci{\sc tsr}~\cite{chi2019complicated}, {\sc t}able2{\sc l}atex~\cite{tabletolatex}, and {\sc p}ub{\sc t}ab{\sc n}et~\cite{pubtabnet} are used exclusively for table structure recognition and table recognition purpose in the literature.  

\subsection{Related Datasets}

\paragraph{\textbf{ICDAR-2013~\cite{gobel2013icdar}:}}

This dataset is one of the most cited datasets for the task of table detection and table structure recognition. It contains 67 {\sc pdf}s which corresponds to 238 page images. Among them, 40 {\sc pdf}s are taken from the {\sc us} Government and 27 {\sc pdf}s from {\sc eu}. Among 238 page images, only 135 page images contain tables, in total 150 tables. This dataset is popularly used to evaluate the algorithms for both table detection and structure recognition task.

\paragraph{\textbf{ICDAR-POD-2017~\cite{icdar_2017}:}}

It focuses on the detection of various graphical objects (e.g., tables, equations, and figures) in the document images. It is created by annotating 2417 document images selected from 1500 scientific papers of CiteSeer. It includes a large variety in page layout - single-column, double-column, multi-column, and a significant variation in the object structure. This dataset is divided into (i) training set consisting of 1600 images, and (ii) test set comprising 817 images. 

\paragraph{\textbf{cTDaR~\cite{icdar_table_2019}:}}

This dataset consists of modern and archival documents with various formats, including document images and born-digital formats such as {\sc pdf}. The images show a great variety of tables from hand-drawn accounting books to stock exchange lists and train timetable, from record books of prisoner lists, tables from printed books, production census, etc. The modern documents consist of scientific journals, forms, financial statements, etc. Annotations correspond to table regions, and cell regions are available. This dataset contains (i) training set - consists of 600 annotated archival and 600 annotated modern document images with table bounding boxes, and (ii) test set - includes 199 annotated archival and 240 annotated modern document images with table bounding boxes. 

\paragraph{\textbf{Marmot~\cite{marmot}:}}

It consists of 2000 Chinese and English pages at the proportion of about 1:1. Chinese pages are selected from over 120 e-Books with diverse subject areas provided by Founder Apabi library. Not more than 15 pages are selected from each Book. English pages are selected over 1500 journal and conference papers published during 1970-2011 from the Citeseer website. The pages show a great variety in layout - one-column and two-column, language type, and table styles. This dataset is used for both table detection and structure recognition tasks.

\paragraph{\textbf{UNLV~\cite{unlv}:}} 

It contains 2889 document images of various categories: technical reports, magazines, business letters, newspapers, etc. Among them, only 427 images include 558 table zones. Annotations for table regions and cell regions are available. This dataset is also used for both table detection and table structure recognition tasks. 

\paragraph{\textbf{DeepFigures~\cite{siegel2018extracting}:}}

Siegel \textit{et al.}~\cite{siegel2018extracting} create this dataset by automatically annotating pages of two large web collections of scientific documents (ar{\sc x}iv and {\sc p}ub{\sc m}ed). This dataset consists of 5.5 million pages with 1.4 million tables and 4.0 million figures. It can be used for detecting tables and figures in the document images.   
\paragraph{\textbf{PubLayNet~\cite{publaynet}:} and \textbf{PubTabNet~\cite{pubtabnet}}}

Pub{\sc l}ay{\sc n}et consists of 360{\sc k} page images with annotation of various layout elements such as text, title, list, figure, and table. It is created by automatically annotating 1 million {\sc p}ub{\sc m}ed Central$^{TM}$ {\sc pdf} articles. It contains 113{\sc k} table regions and 126{\sc k} figure regions in total. It is mainly used for layout analysis purposes. However, it can also be used for table and figure detection tasks.  
{\sc p}ub{\sc t}ab{\sc n}et is the largest dataset for the table recognition task. It consists of 568{\sc k} images of heterogeneous tables extracted from scientific articles (in {\sc pdf} format). Ground truth corresponds to each table image represents structure information and text content of each cell in {\sc html} format.   

\paragraph{\textbf{TableBank~\cite{tablebank}}:}

This dataset contains 417{\sc k} high-quality documents with tables. This dataset is created by downloading LaTex source code from the year 2014 to the year 2018 through bulk data access in ar{\sc x}iv. It consists of 163{\sc k} word document, 253{\sc k} LaTex document and in total 417{\sc k} document images with annotated table regions. Structure level annotation is available for 57{\sc k} word documents, 88{\sc k} LaTex documents and in total 145{\sc k} document images.

\subsection{Related Work in Graphical Object Detection}

Localizing graphical objects (e.g., tables, figures, mathematical equations, logos, signatures, etc.) is the primary step for understanding any documents. Recent advances in object detection in natural scene images using deep learning inspire researchers~\cite{gilani2017table,schreiber2017deepdesrt,kavasidis2019saliency,Siddiqui2018DeCNTDD,melinda_2019,li_2019,yilun_2019,sun_2019,ranajit_2019} to develop deep learning based algorithms for detecting graphical objects in documents. In this regards, various researchers like Gilani \textit{et al.}~\cite{gilani2017table}, Schreiber \textit{et al.}~\cite{schreiber2017deepdesrt}, Siddiqui \textit{et al.}~\cite{Siddiqui2018DeCNTDD} and Sun \textit{et al.}~\cite{sun_2019} employ {\sc f}aster {\sc r-cnn}~\cite{ren2015faster} model for detecting table in documents. In~\cite{yilun_2019}, the authors use {\sc yolo}~\cite{redmon2016you} to detect tables in documents. Li \textit{et al.}~\cite{li_2019} use Generative Adversarial Networks ({\sc gan})~\cite{goodfellow2014generative} to extract layout feature to improve table detection accuracy.

Few researchers~\cite{kavasidis2019saliency,ranajit_2019} focus on detection of various kinds of graphical objects (not just tables). Kavasidis \textit{et al.}~\cite{kavasidis2019saliency} propose saliency based technique to detect three types of graphical objects --- table, figure and mathematical equation. In~\cite{ranajit_2019}, {\sc m}ask {\sc r-cnn} is explored to detect various graphical objects.        

\subsection{Baseline Methods for Graphical Object Detection} 

We use {\sc f}aster {\sc r-cnn}~\cite{ren2015faster} and {\sc m}ask {\sc r-cnn}~\cite{he2017mask} as baselines for detecting graphical objects in annual reports. We use publicly available implementations of {\sc f}aster {\sc r-cnn}~\cite{jjfaster2rcnn} and {\sc m}ask {\sc r-cnn}~\cite{matterport_maskrcnn_2017} for this experiment. We train both the models with the images of training set of {\sc iiit-ar}-13{\sc k} dataset for the empirical studies.

\paragraph{\textbf{Faster R-CNN:}}

The model is implemented using PyTorch; trained and evaluated on {\sc nvidia} {\sc titan} {\sc x} {\sc gpu} with 12{\sc gb} memory with batch size of 4. The input images are resized to a fixed size of 800 $\times$ 1024 by preserving original aspect ratio. We use the pre-trained ResNet-101~\cite{he2016deep} backbone on {\sc ms}-{\sc coco}~\cite{lin2014microsoft} dataset. We use five different anchor scales (i.e., 32, 64, 128, 256, and 512) and five anchor ratios (i.e., 1, 2, 3, 4, and 5) so that the region proposals can cover almost every part of the image irrespective of the image size. We use stochastic gradient descent ({\sc sgd}) as an optimizer with initial learning rate = 0.001 and the learning rate decays after every 5 epochs and it is equal to 0.1 times of the previous value.

\paragraph{\textbf{Mask R-CNN:}} 

Every input image is rescaled to a fixed size of 800 $\times$ 1024 preserving the original aspect ratio. We use the pre-trained ResNet-101~\cite{he2016deep} backbone on {\sc ms}-{\sc coco}~\cite{lin2014microsoft} dataset. We use Tensorflow/Keras for the implementation, and train and evaluate on {\sc nvidia} {\sc titan} {\sc x} {\sc gpu} that has 12{\sc gb} memory with a batch size of 1. We further use 0.5, 1, and 2 as the anchor scale values and anchor box sizes of 32, 64, 128, 256, and 512. Further, we train a total of 80 epochs. We train all {\sc fpn} and subsequent layers for first 20 epochs, the next 20 epochs for training {\sc fpn} + last 4 layers of ResNet-101; and last 40 epochs for training all layers of the model. During the training process, we use 0.001 as the learning rate, 0.9 as the momentum and 0.0001 as the weight decay.

\begin{figure}[ht!]
\centerline{
\tcbox[sharp corners, size = tight, boxrule=0.4mm, colframe=black, colback=white]{
\epsfig{figure=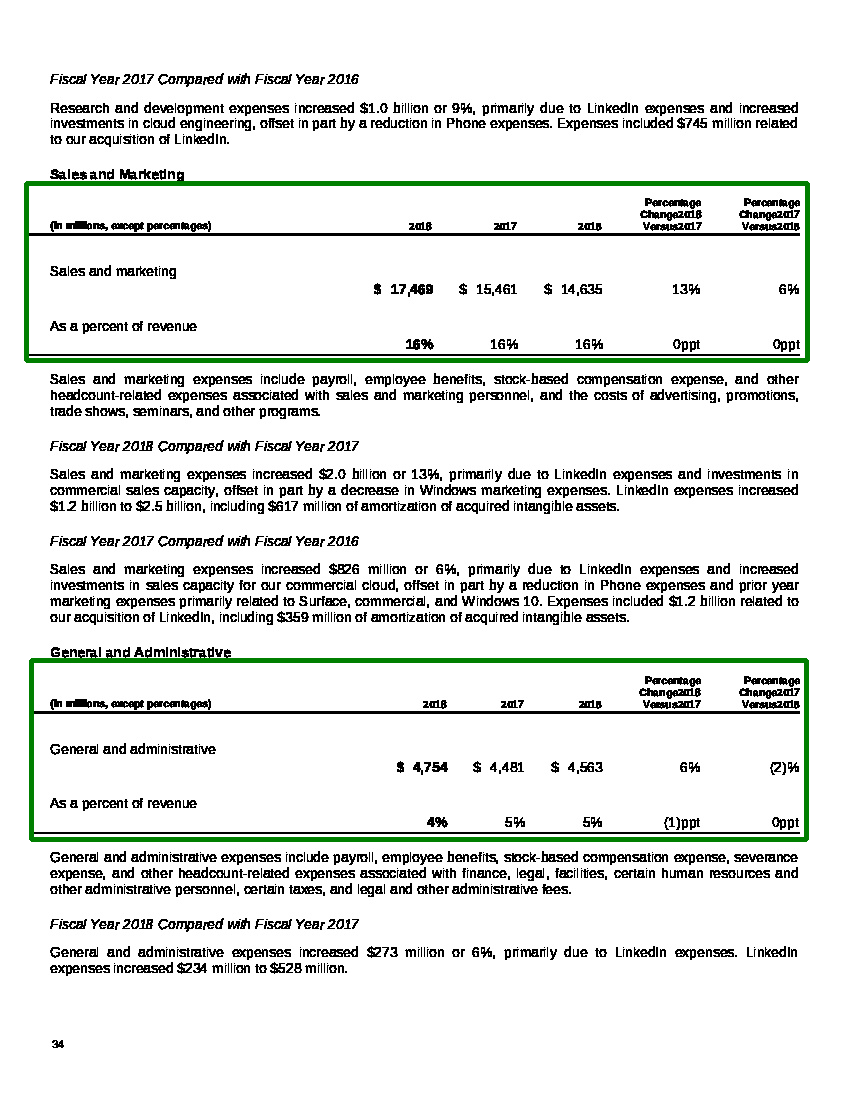, width=0.23\textwidth,height=0.2\textwidth}}
\hspace{-0.001\textwidth}
\tcbox[sharp corners, size = tight, boxrule=0.4mm, colframe=black, colback=white]{
\epsfig{figure=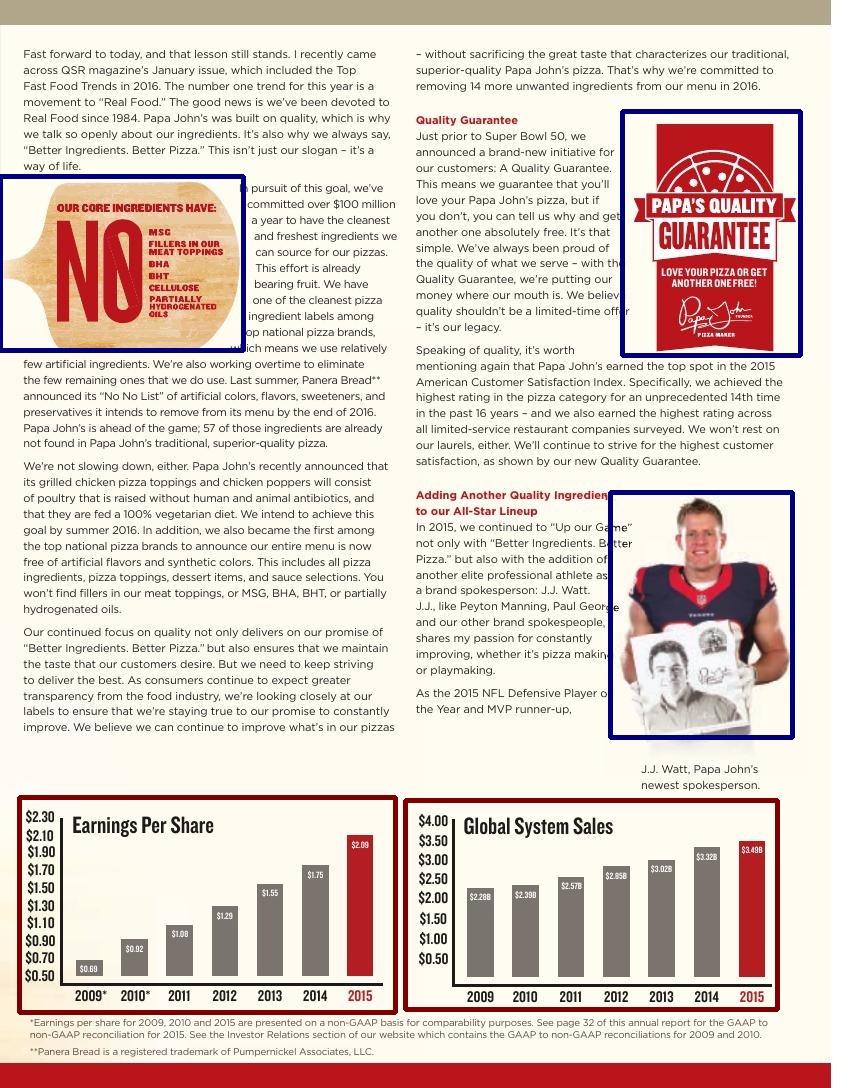, width=0.23\textwidth,height=0.2\textwidth}}
\hspace{-0.001\textwidth}
\tcbox[sharp corners, size = tight, boxrule=0.4mm, colframe=black, colback=white]{
\epsfig{figure=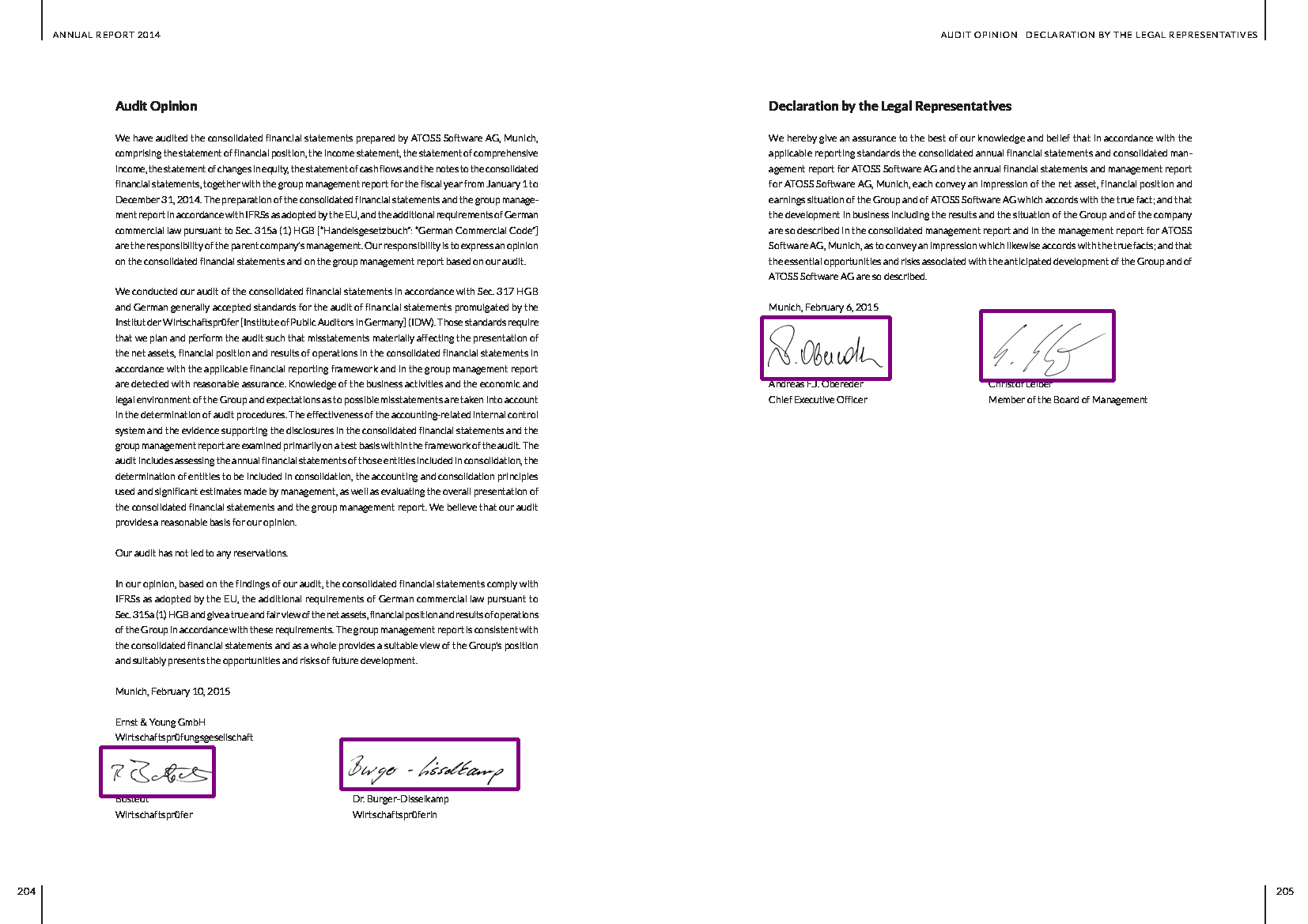, width=0.23\textwidth,height=0.2\textwidth}}
\hspace{-0.001\textwidth}
\tcbox[sharp corners, size = tight, boxrule=0.4mm, colframe=black, colback=white]{
\epsfig{figure=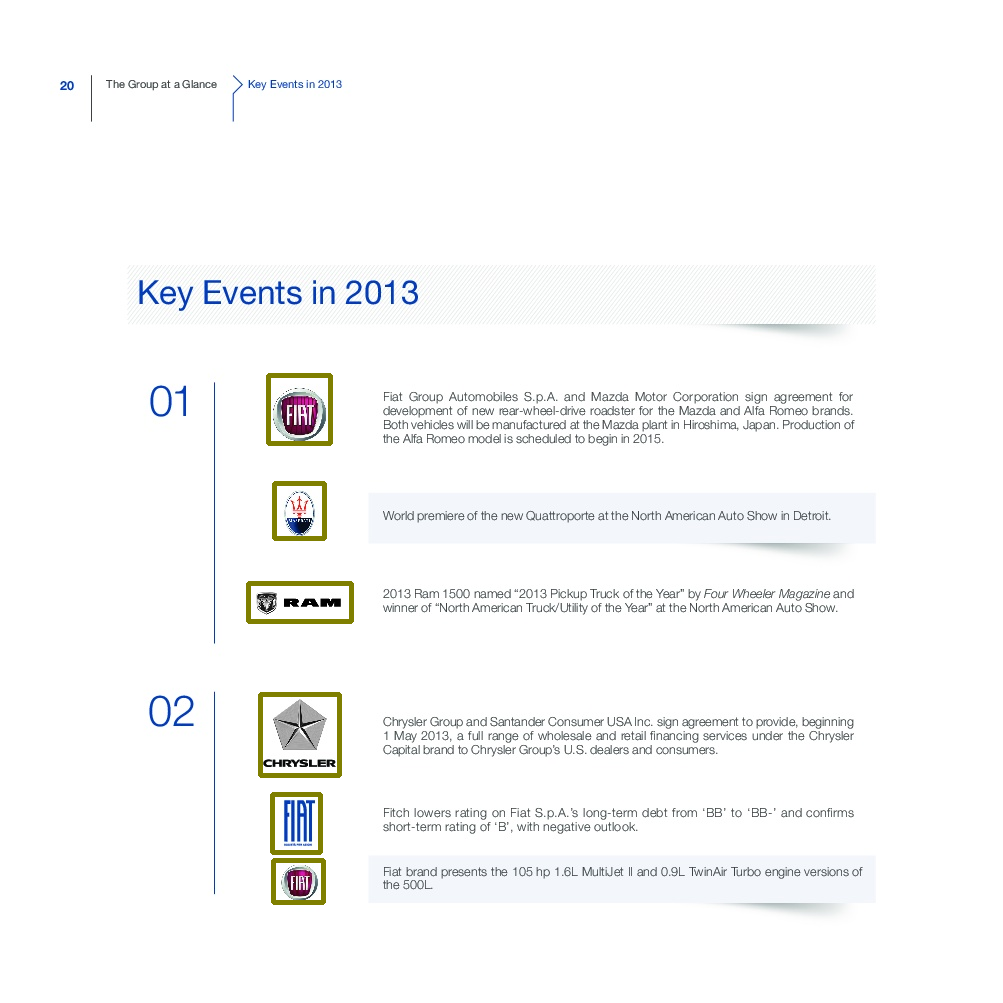, width=0.23\textwidth,height=0.2\textwidth}}}
\caption{Sample annotated document images of {\sc iiit-ar}-13{\sc k} dataset. \textbf{Dark Green:} indicates ground truth bounding box of table, \textbf{Dark Red:} indicates ground truth bounding box of figure, \textbf{Dark Blue:} indicates ground truth bounding box of natural image, \textbf{Dark Yellow:} indicates ground truth bounding box of logo and \textbf{Dark Pink:} indicates ground truth bounding box of signature. \label{fig_sample_annotation}}
\end{figure}

\section{IIIT-AR-13K Dataset}
\vspace{-0.01\textwidth}
\subsection{Details of the Dataset}

For detecting various types of graphical objects (e.g., table, figure, natural image, logo, and signature) in the business documents, we generate a new dataset, named {\sc iiit-ar}-13{\sc k}. The newly generated dataset consists of 13{\sc k} pages of publicly available annual reports. Annual reports in English and non-English languages (e.g., French, Japanese, Russian, etc.) of multiple (more than ten) years of twenty-nine different companies. Annual reports contain various types of graphical objects such as tables, various types of graphs (e.g., bar chart, pie chart, line plot, etc.), images, companies' logos, signatures, stamps, sketches, etc. However, we only consider five categories of graphical objects, e.g., table, figure (including both graphs and sketches), natural image (including images), logo, and signature. 

\begin{table}[ht!]
\begin{center}
\begin{tabular}{|l | r| r| r| r|} \hline
\textbf{Object Category} & \multicolumn{4}{|c| }{\textbf{IIIT-AR-13K}}  \\ \cline{2-5}
 &\textbf{Training} &\textbf{Validation} &\textbf{Test} &\textbf{Total} \\ \hline
\textbf{Page} &9333 $\approx$ 9K &1955 $\approx$ 2K &2120 $\approx$ 2K &13415 $\approx$ 13K \\ \hline
\textbf{Table} & 11163 $\approx$ 11K &2222 $\approx$ 2K &2596 $\approx$ 3K &15981 $\approx$ 16K \\
\textbf{Figure} & 2004 $\approx$ 2K &481 $\approx$ 0.4K & 463 $\approx$ 0.4K &2948 $\approx$ 3K \\
\textbf{Natural Image} &1987 $\approx$ 2K &438 $\approx$ 0.4K &455 $\approx$ 0.4K &2880 $\approx$ 3K \\
\textbf{Logo} &379 $\approx$ 0.3K &67 $\approx$ 0.06K &135 $\approx$ 0.1K &581 $\approx$ 0.5K \\
\textbf{Signature} &420 $\approx$ 0.4K & 108 $\approx$ 0.9K &92 $\approx$ 0.09K & 620 $\approx$ 0.6K \\ \hline
\end{tabular}
\end{center}
\caption{Statistics of our newly generated {\sc iiit-ar-13k} dataset.\label{table_ar_dataset_statistics}} \end{table}

Annual reports in English and other languages of more than ten years of twenty-nine different companies are selected to increase diversity with respect to the language, layout, and each category of graphical objects. We manually annotate the bounding boxes of each type of graphical object. Figure~\ref{fig_sample_annotation} shows some sample bounding box annotated images. Finally, we have 13{\sc k} annotated pages with 16{\sc k} tables, 3{\sc k} figures, 3{\sc k} natural images, 0.5{\sc k} logos, and 0.6{\sc k} signatures. The dataset is divided into (i) training set consisting of 9{\sc k} document images, (ii) validation set containing 2{\sc k} document images, and (iii) test set consists of 2{\sc k} document images. For every company, we randomly choose 70\%, 15\%, and remaining 15\% of the total pages as training, validation, and test, respectively. Table~\ref{table_ar_dataset_statistics} shows the statistics of the newly generated dataset.

\subsection{Comparison with the Existing Datasets}

Table~\ref{table_existing_dataset_statistics} presents the comparison of our dataset with the existing datasets. From the table, it is clear that with respect to label space, all the existing datasets (except {\sc icdar-pod-2017}, {\sc d}eep{\sc f}igures, and {\sc p}ub{\sc l}ay{\sc n}et) containing only one object category, i.e., table, are subsets of our newly generated {\sc iiit-ar-13k} dataset (containing five object categories). Most of the existing datasets (except {\sc icdar-2013}, c{\sc td}a{\sc r}, and {\sc unlv}) consist of only scientific research articles, handwritten documents, and e-books. The newly generated dataset includes annual reports, one specific type of business document. It is the largest among the existing datasets where ground truths are annotated manually for the detection task. It consists of a large variety of pages of annual reports compared to scientific articles of existing datasets.

\begin{table}[ht!]
\begin{center}
\begin{tabular}{|l | l| r| c| c| c| c| r|} \hline
\textbf{Dataset} & \textbf{Category Label}  &  \textbf{\#Images} & \multicolumn{4}{|c| }{\textbf{Task}} &\textbf{\#Tables} \\ \cline{4-7}
 & & &\textbf{POD}&\textbf{TD}&\textbf{TSR}&\textbf{TR} & \\ \hline
{\sc icdar}-2013~\cite{gobel2013icdar} &1: T & 238 & \checkmark &\checkmark &\checkmark &\checkmark &150 \\ \hline
{\sc icdar-pod}-2017~\cite{icdar_2017}&3: T, F, E & 2417 &\checkmark& \checkmark&$\times$ &$\times$ &1020 \\ \hline
c{\sc td}a{\sc r}~\cite{icdar_table_2019}&1: T &2K &\checkmark & \checkmark &\checkmark &\checkmark &3.6K \\ \hline
{\sc m}armot~\cite{marmot} &1: T &2K &\checkmark& \checkmark&$\times$ &$\times$ &958 \\ \hline
{\sc unlv}~\cite{unlv} &1: T &427 &\checkmark& \checkmark&\checkmark &\checkmark &558 \\ \hline
\textbf{IIIT-AR-13k} &\textbf{5:} \textbf{T}, \textbf{F}, \textbf{NI}, \textbf{L}, \textbf{S} &\textbf{13K} &\checkmark & \checkmark &$\times$&$\times$ &\textbf{16K} \\ \hline 
{\sc d}eep{\sc f}igures~\cite{siegel2018extracting}\footnotemark[2] &2: T, F& 5.5M &\checkmark& \checkmark&$\times$ &$\times$&1.4M \\ \hline
{\sc p}ub{\sc l}ay{\sc n}et~\cite{publaynet}\footnotemark[2] &5: T, F, TL, TT, LT &360K &\checkmark& \checkmark&$\times$ &$\times$ &113K \\ \hline
{\sc t}able{\sc b}ank~\cite{tablebank}\footnotemark[2]&1: T &417K &\checkmark &\checkmark &$\times$ &$\times$ &417K \\ \cline{4-8}
 & &  &$\times$ &$\times$ &\checkmark&$\times$ &145K \\ \hline
{\sc s}ci{\sc tsr}~\cite{chi2019complicated}\footnotemark[3] &1: T &- &$\times$ &$\times$ &\checkmark &\checkmark &15K \\ \hline
{\sc t}able2{\sc l}atex~\cite{tabletolatex}\footnotemark[3] &1: T & - &$\times$ &$\times$ &\checkmark &\checkmark &450K \\ \hline 
{\sc p}ub{\sc t}ab{\sc n}et~\cite{pubtabnet}\footnotemark[3] &1: T &- &$\times$ &$\times$ &\checkmark &\checkmark &568K \\ \hline 
\end{tabular}
\end{center}
\caption{Statistics of existing datasets along with newly generated dataset for graphical object detection task in document images. \textbf{T:} indicates table. \textbf{F:} indicates figure. \textbf{E:} indicates equation. \textbf{NI:} indicates natural image. \textbf{L:} indicates logo. \textbf{S:} indicates signature. \textbf{TL:} indicates title. \textbf{TT:} indicates text. \textbf{LT:} indicates list. \textbf{POD:} indicates page object detection. \textbf{TD:} indicates table detection. \textbf{TSR:} indicates table structure recognition. \textbf{TR:} indicates table recognition.\label{table_existing_dataset_statistics}} \end{table}
\footnotetext[2]{Ground truth bounding boxes are annotated automatically.}
\footnotetext[3]{Dataset is dedicated only for table structure recognition and table recognition.}

\begin{figure}[ht!]
\centerline{
\tcbox[sharp corners, size = tight, boxrule=0.4mm, colframe=black, colback=white]{
\epsfig{figure=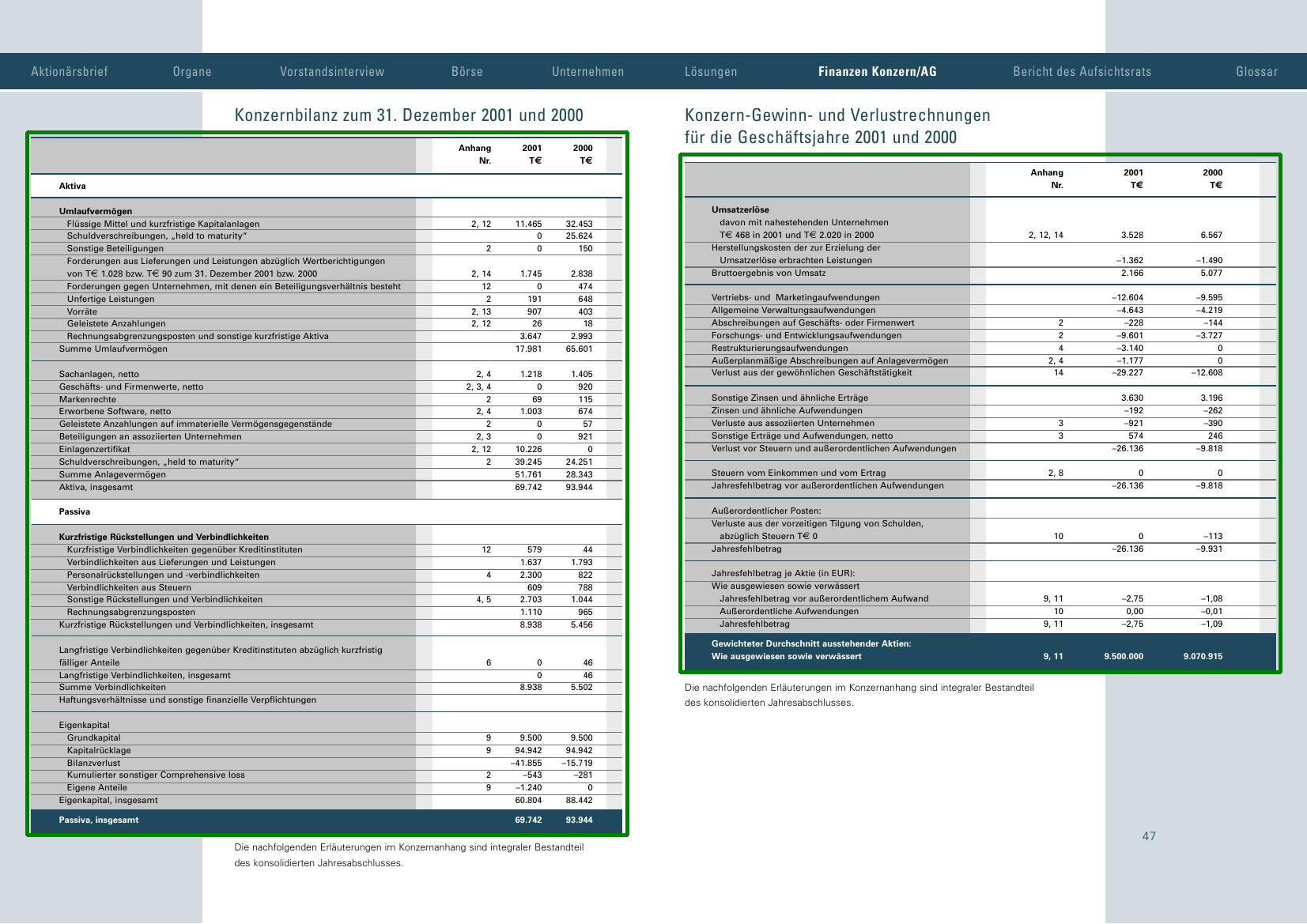, width=0.23\textwidth,height=0.16\textwidth}}
\hspace{-0.002\textwidth}
\tcbox[sharp corners, size = tight, boxrule=0.4mm, colframe=black, colback=white]{
\epsfig{figure=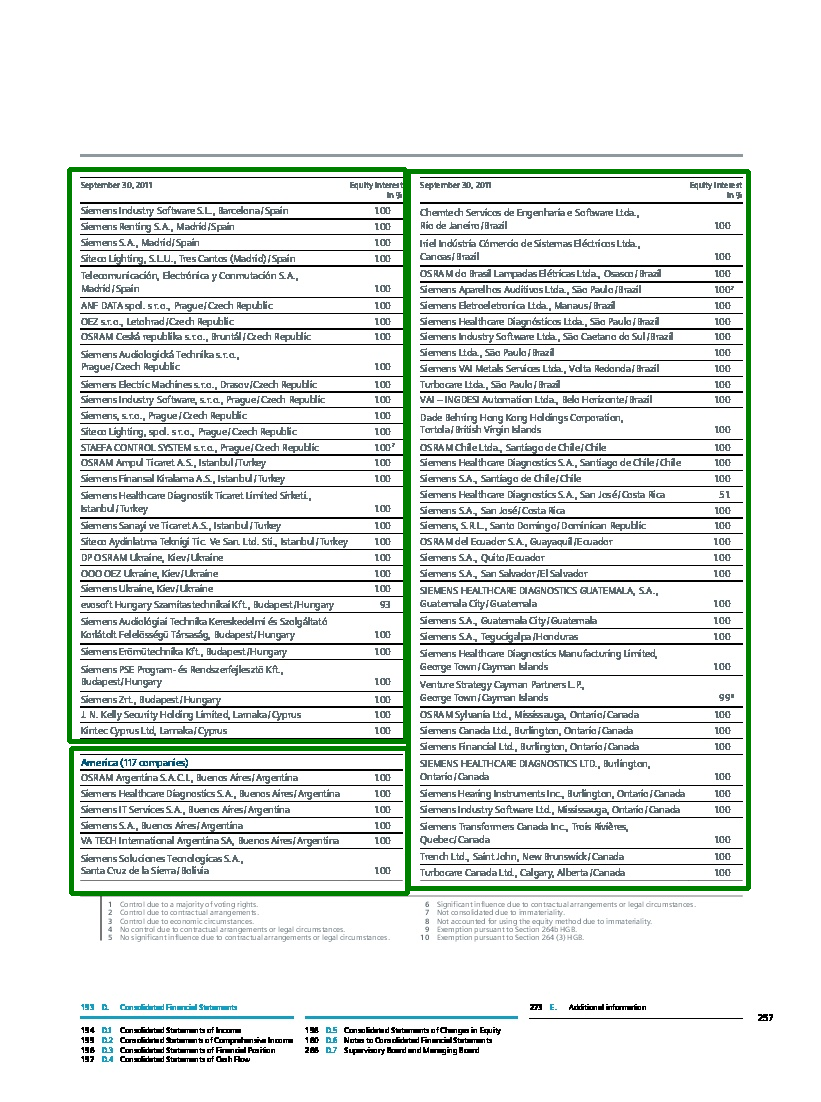, width=0.23\textwidth,height=0.16\textwidth}}
\hspace{-0.002\textwidth}
\tcbox[sharp corners, size = tight, boxrule=0.4mm, colframe=black, colback=white]{
\epsfig{figure=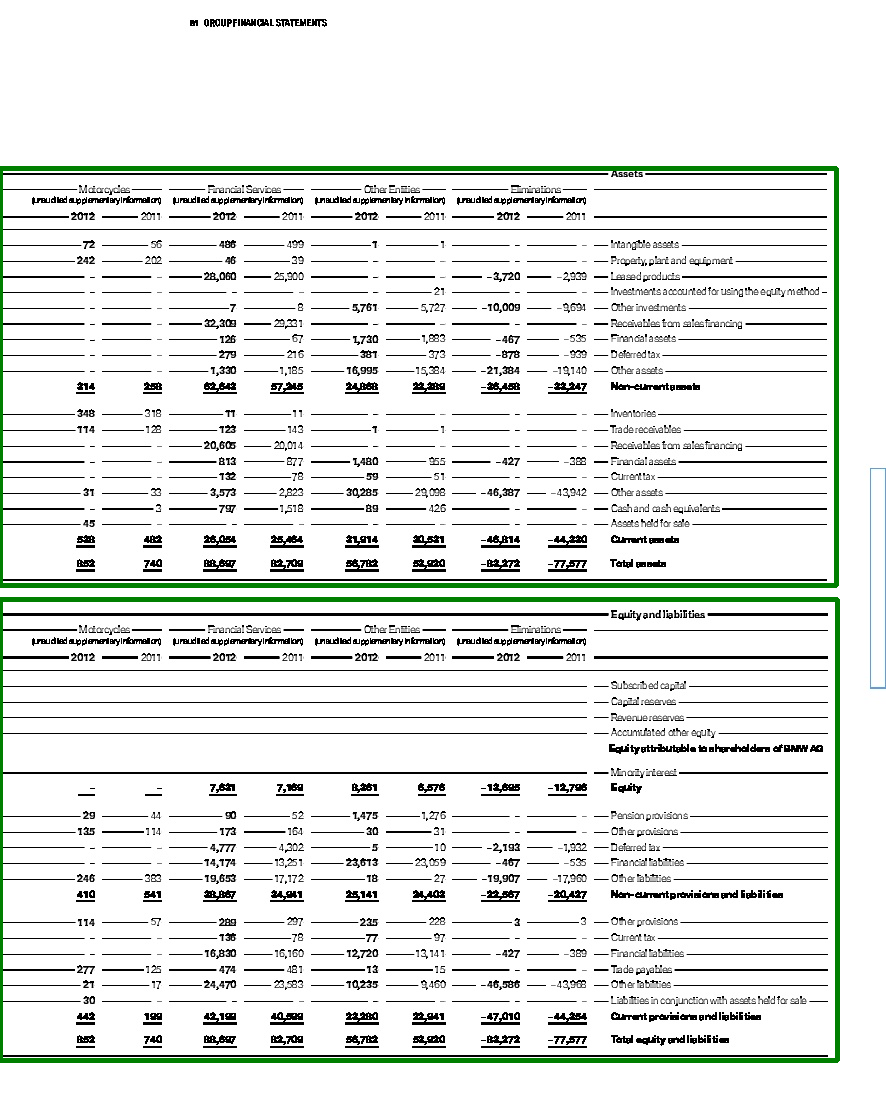, width=0.23\textwidth,height=0.16\textwidth}}
\hspace{-0.002\textwidth}
\tcbox[sharp corners, size = tight, boxrule=0.4mm, colframe=black, colback=white]{
\epsfig{figure=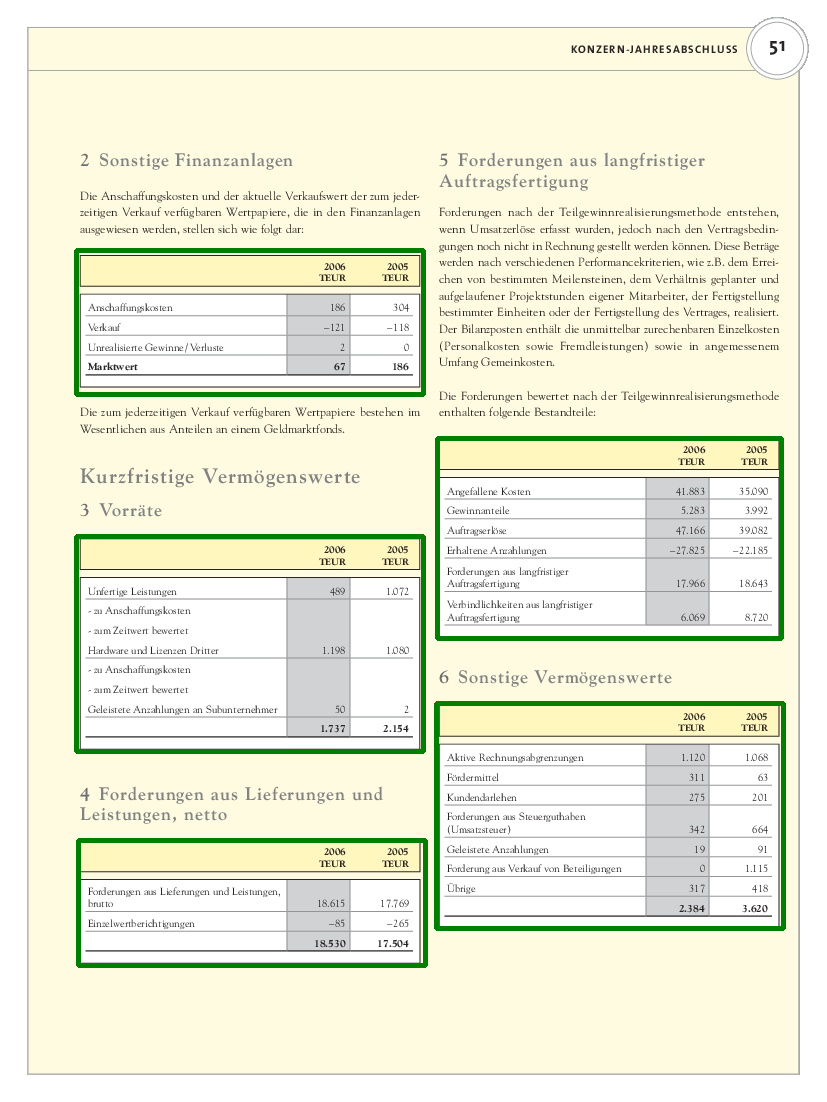, width=0.23\textwidth,height=0.16\textwidth}}}
\vspace{-0.03\textwidth}
\centerline{
\tcbox[sharp corners, size = tight, boxrule=0.4mm, colframe=black, colback=white]{
\epsfig{figure=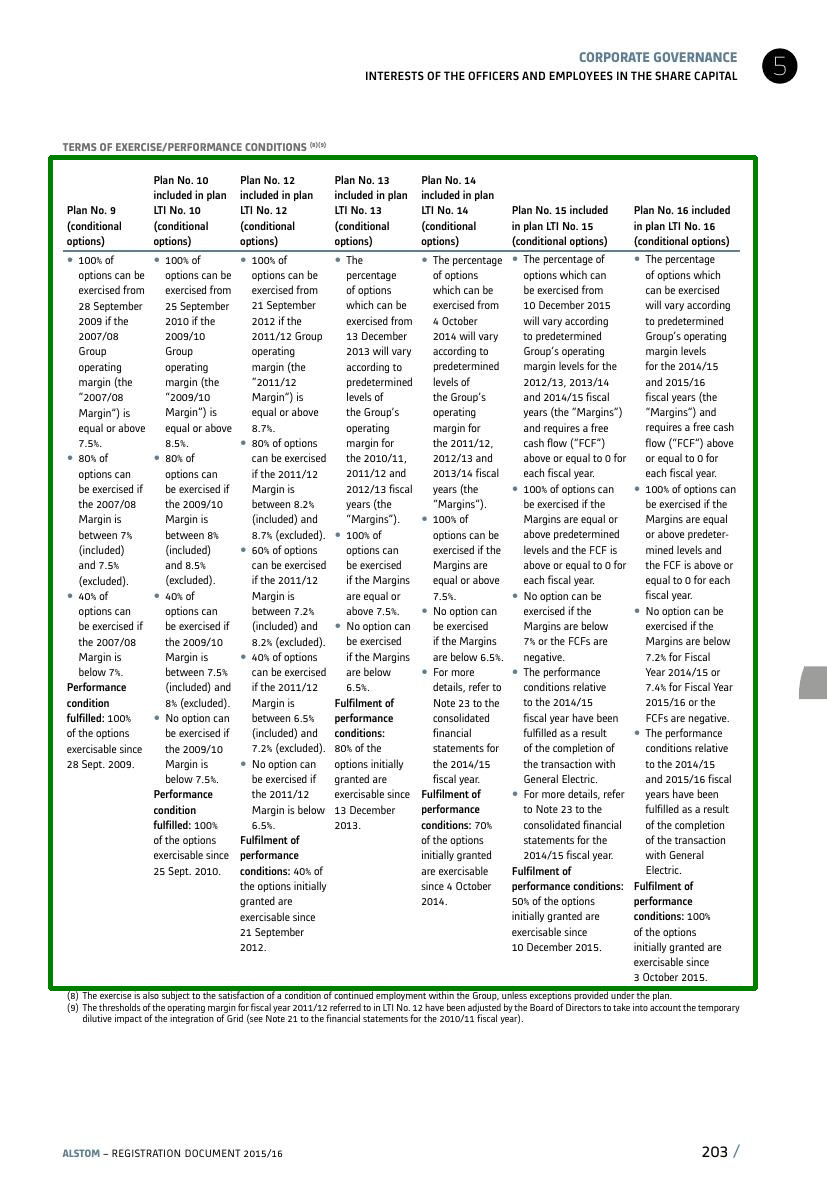, width=0.23\textwidth,height=0.16\textwidth}}
\hspace{-0.002\textwidth}
\tcbox[sharp corners, size = tight, boxrule=0.4mm, colframe=black, colback=white]{
\epsfig{figure=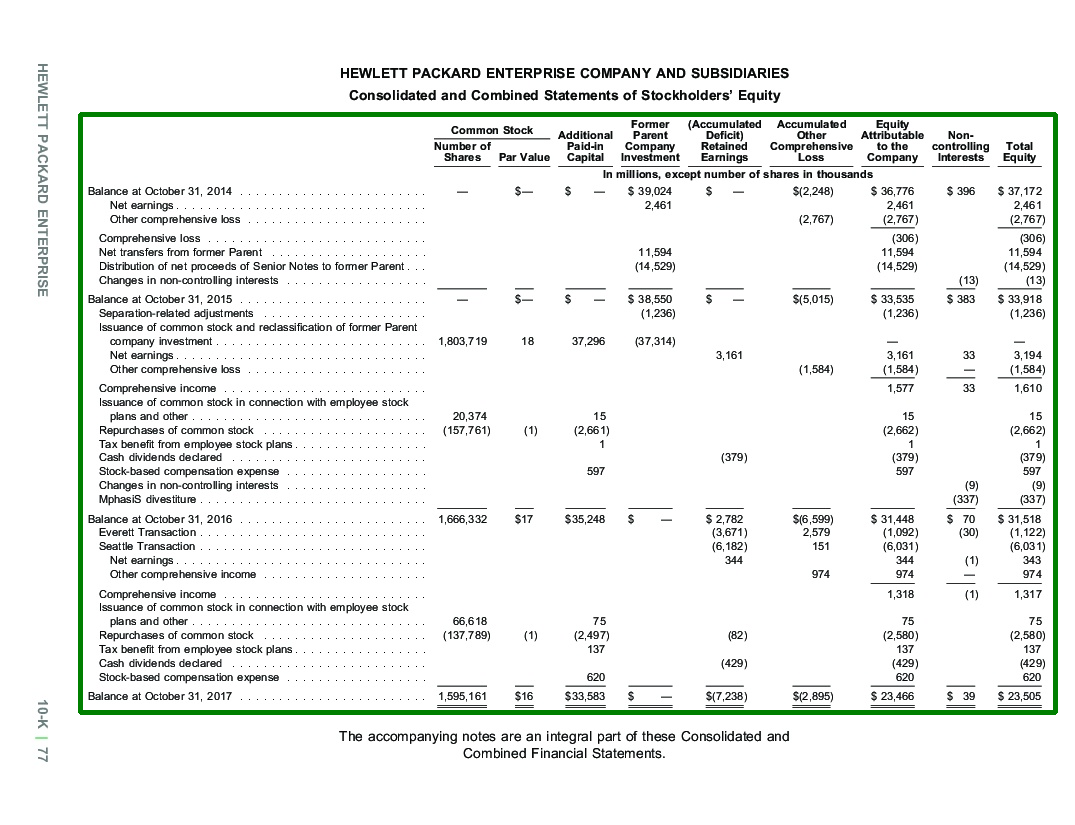, width=0.23\textwidth,height=0.16\textwidth}}
\hspace{-0.002\textwidth}
\tcbox[sharp corners, size = tight, boxrule=0.4mm, colframe=black, colback=white]{
\epsfig{figure=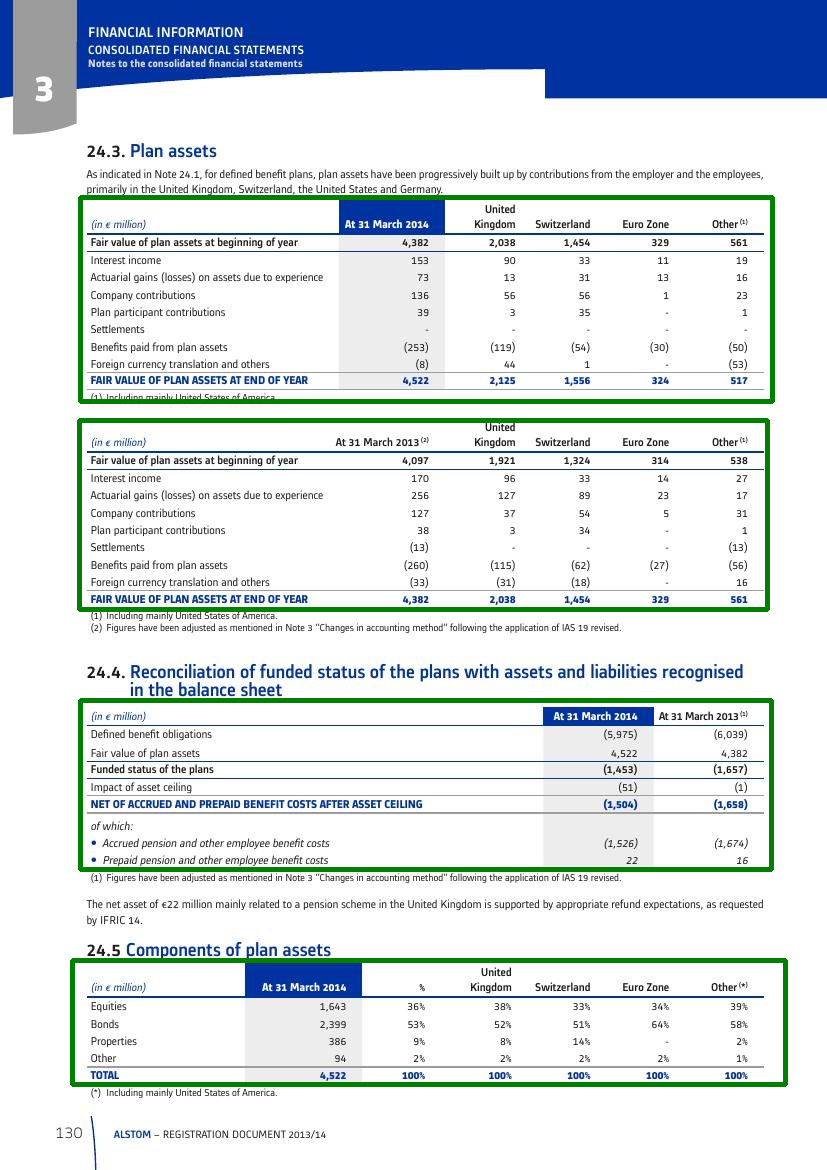, width=0.23\textwidth,height=0.16\textwidth}}
\hspace{-0.002\textwidth}
\tcbox[sharp corners, size = tight, boxrule=0.4mm, colframe=black, colback=white]{
\epsfig{figure=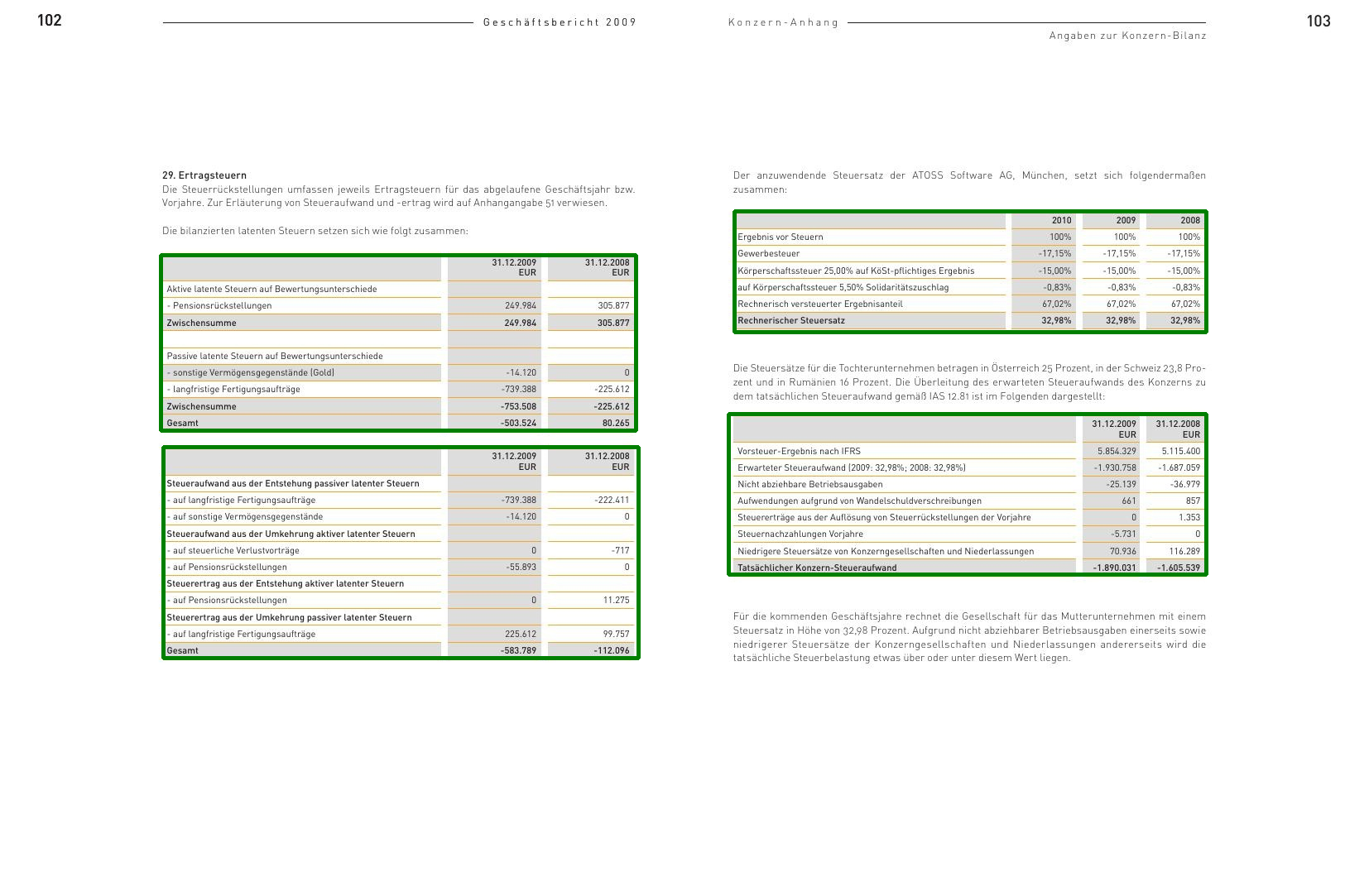, width=0.23\textwidth,height=0.16\textwidth}}}
\vspace{-0.03\textwidth}
\centerline{
\tcbox[sharp corners, size = tight, boxrule=0.4mm, colframe=black, colback=white]{
\epsfig{figure=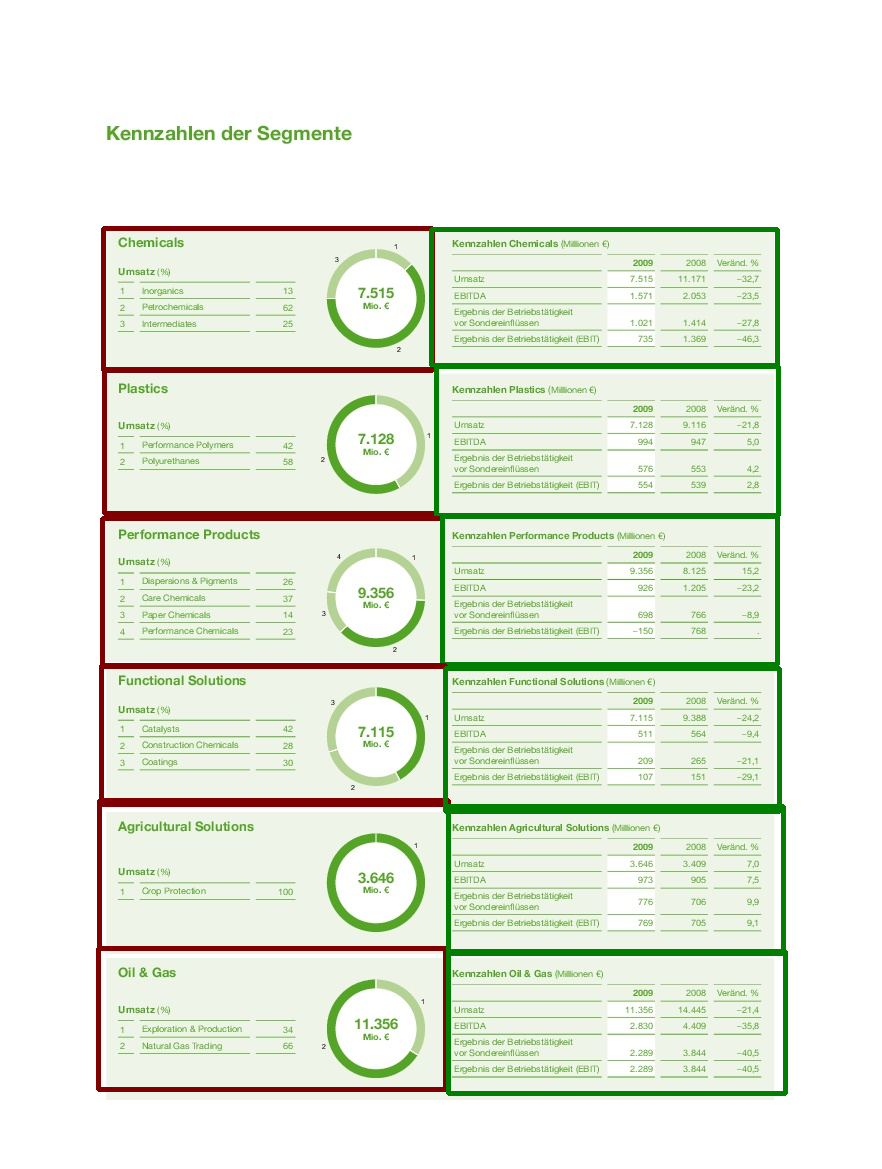, width=0.23\textwidth,height=0.16\textwidth}}
\hspace{-0.002\textwidth}
\tcbox[sharp corners, size = tight, boxrule=0.4mm, colframe=black, colback=white]{
\epsfig{figure=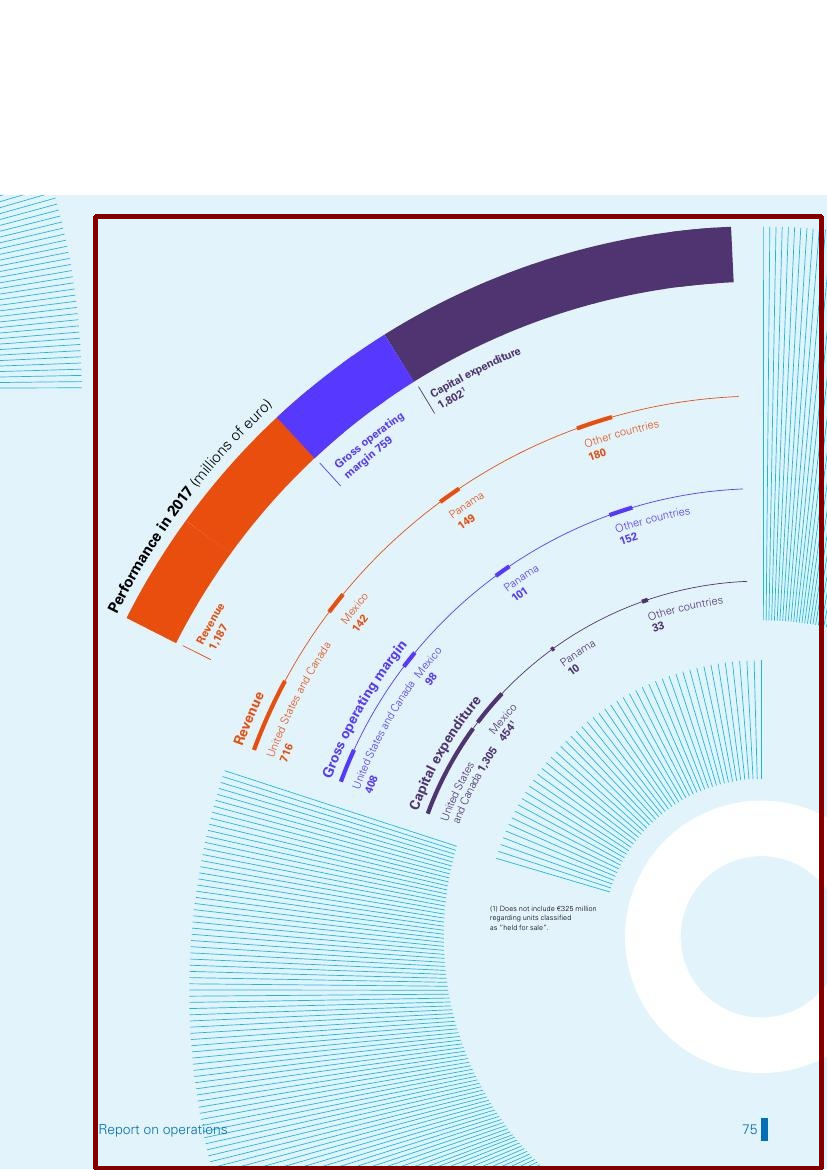, width=0.23\textwidth,height=0.16\textwidth}}
\hspace{-0.002\textwidth}
\tcbox[sharp corners, size = tight, boxrule=0.4mm, colframe=black, colback=white]{
\epsfig{figure=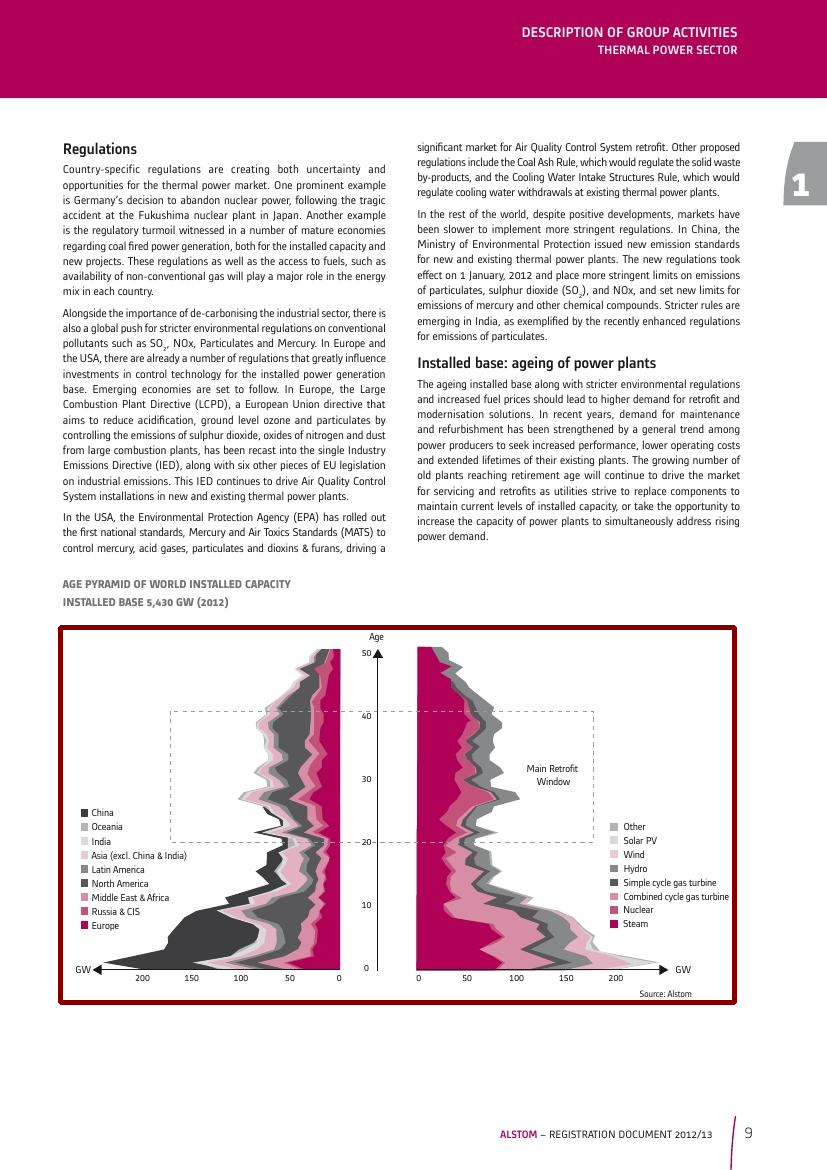, width=0.23\textwidth,height=0.16\textwidth}}
\hspace{-0.002\textwidth}
\tcbox[sharp corners, size = tight, boxrule=0.4mm, colframe=black, colback=white]{
\epsfig{figure=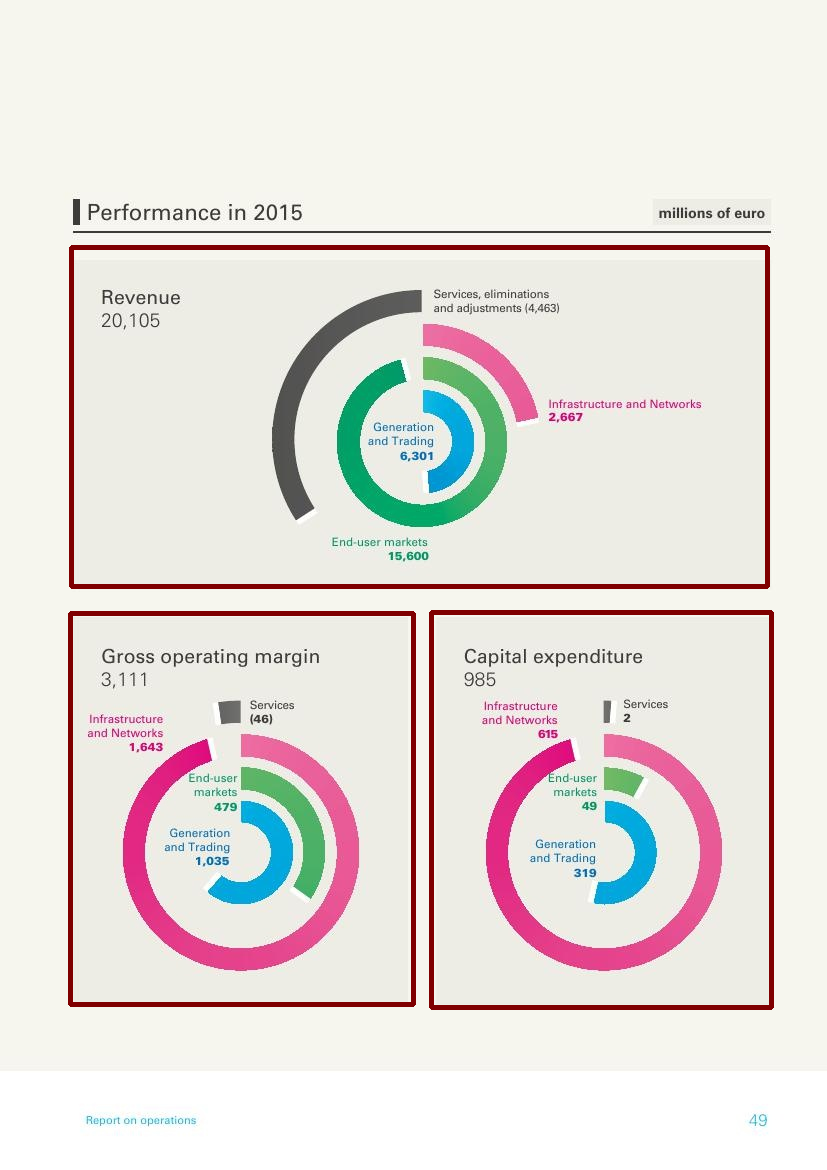, width=0.23\textwidth,height=0.16\textwidth}}}
\vspace{-0.03\textwidth}
\centerline{
\tcbox[sharp corners, size = tight, boxrule=0.4mm, colframe=black, colback=white]{
\epsfig{figure=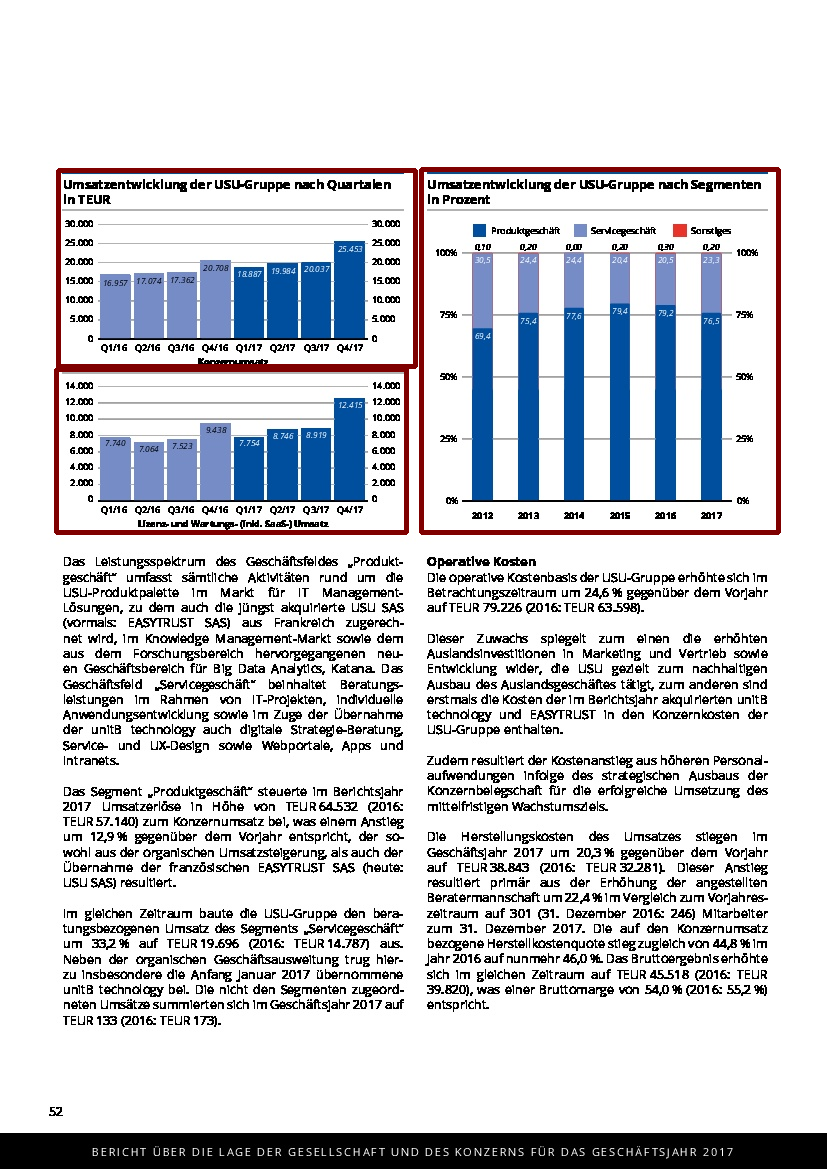, width=0.23\textwidth,height=0.16\textwidth}}
\hspace{-0.002\textwidth}
\tcbox[sharp corners, size = tight, boxrule=0.4mm, colframe=black, colback=white]{
\epsfig{figure=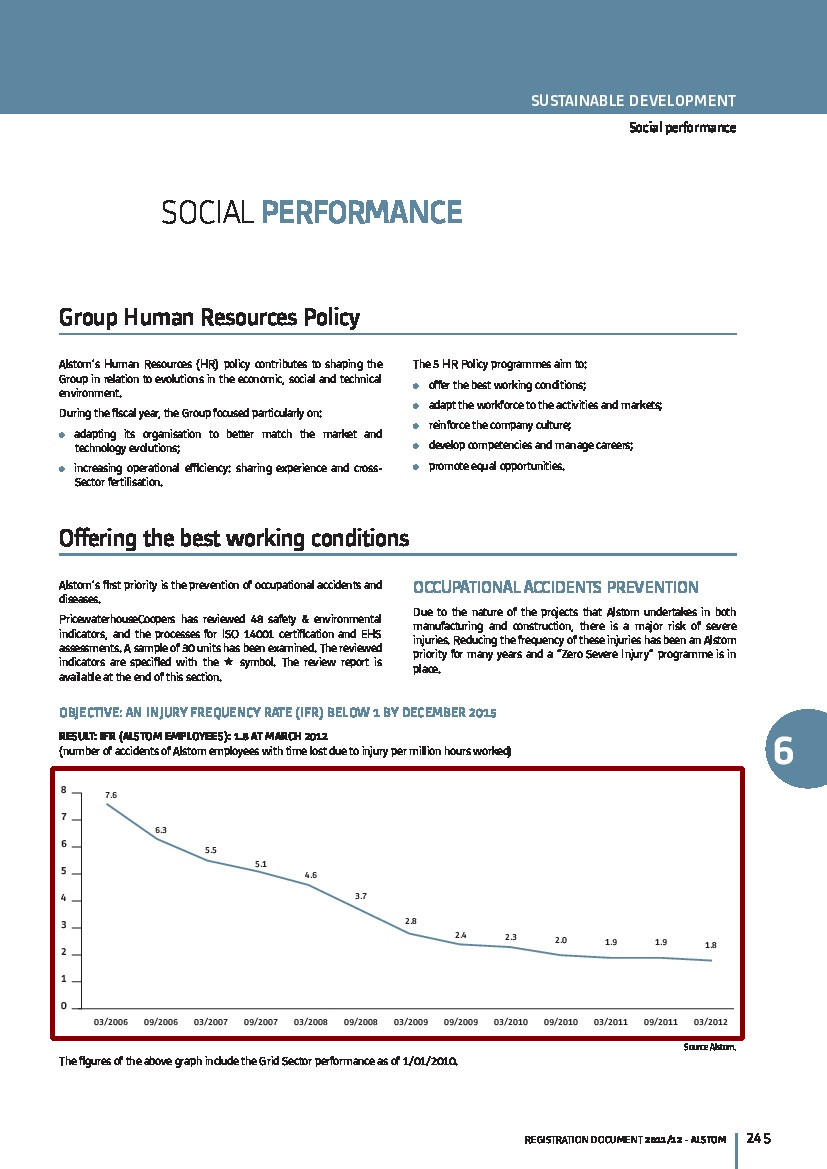, width=0.23\textwidth,height=0.16\textwidth}}
\hspace{-0.002\textwidth}
\tcbox[sharp corners, size = tight, boxrule=0.4mm, colframe=black, colback=white]{
\epsfig{figure=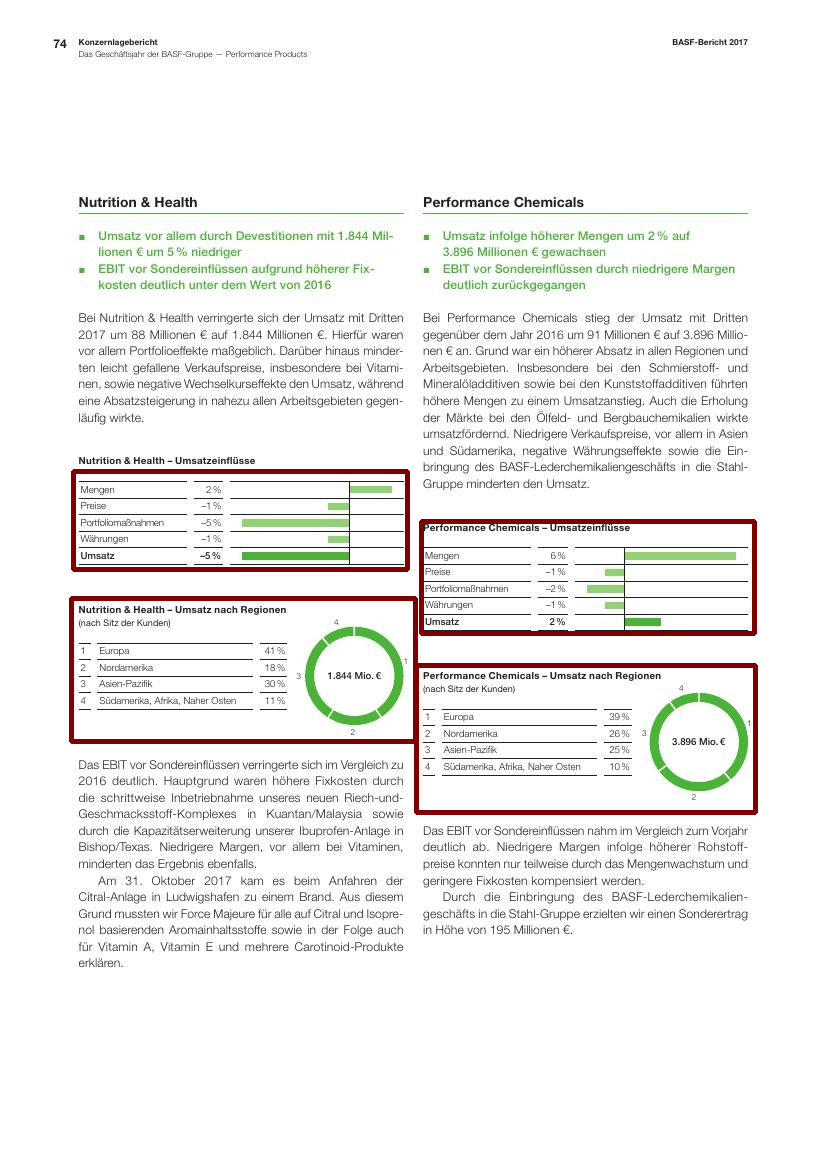, width=0.23\textwidth,height=0.16\textwidth}}
\hspace{-0.002\textwidth}
\tcbox[sharp corners, size = tight, boxrule=0.4mm, colframe=black, colback=white]{
\epsfig{figure=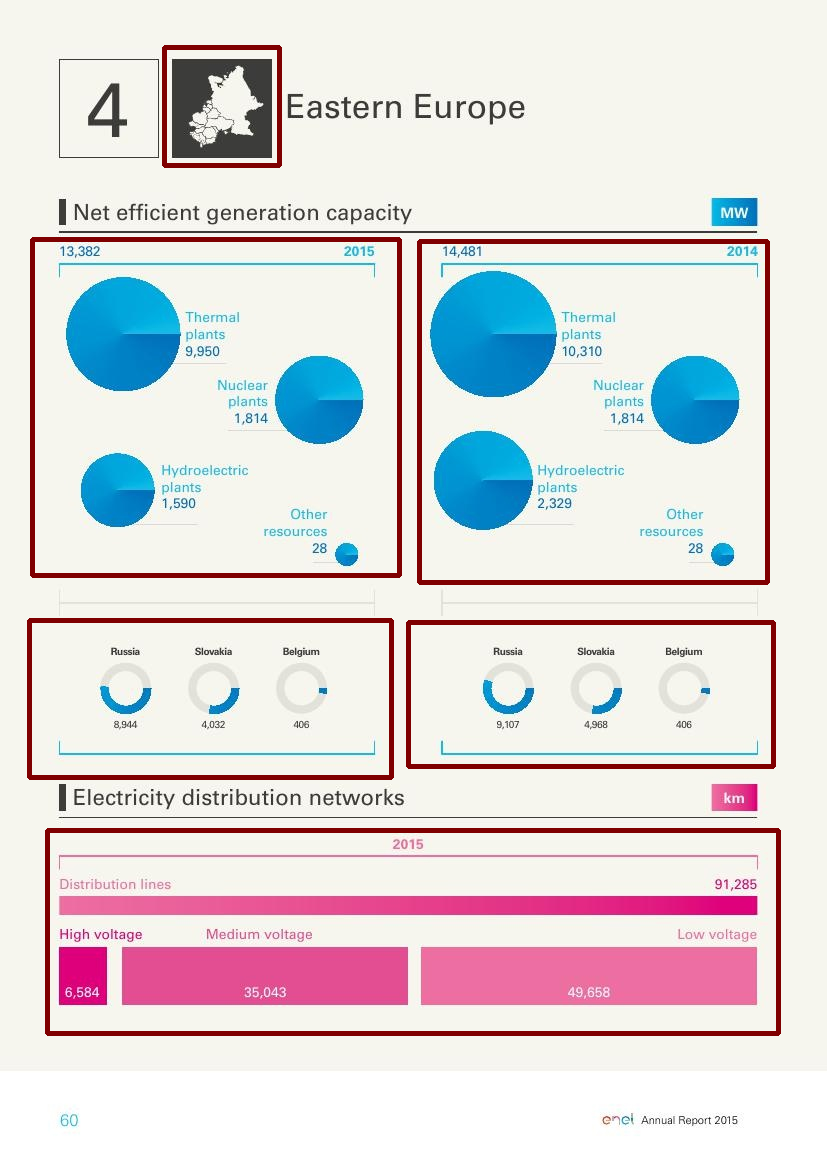, width=0.23\textwidth,height=0.16\textwidth}}}
\caption{Sample annotated images with large variation in tables and figures. \label{fig_table_variation}}
\end{figure}
\vspace{-0.1\textwidth}
\subsection{Diversity in Object Category}

We select annual reports (more than ten years) of twenty-nine different companies for creating this dataset. Due to the consideration of annual reports of several years of various companies, there is a significant variation in layouts (e.g., single-column, double-column, and triple-column) and graphical objects like tables, figures, natural images, logos, and signatures. The selected documents are heterogeneous. Heterogeneity increases the variability within each object category and also increases the similarity between object categories. A significant variation in layout structure, the similarity between object categories, and diversity within object categories make this dataset more complex for graphical detection tasks. Figure~\ref{fig_table_variation} illustrates the significant variations in table structures and figures in the newly generated dataset.

\vspace{-0.01\textwidth}
\subsection{Performance for Detection using Baseline Methods}

Quantitative results of detection on the validation and test sets of {\sc iiit-ar-13k} dataset using baseline approaches - {\sc f}aster {\sc r-cnn} and {\sc m}ask {\sc r-cnn} are summarized in Table~\ref{table_quantitative_result}. From the table, it is observed that {\sc m}ask {\sc r-cnn} produces better results than {\sc f}aster {\sc r-cnn}. Among all the object categories, both the baselines obtain the best results for table and worst results for logo. This is because of highly imbalanced training set (11{\sc k} tables and 0.3{\sc k} logos). 

\begin{table}
\addtolength{\tabcolsep}{-1.0pt}
\begin{center}
\begin{tabular}{|l|l| l| l| l |l| l| l| l|l|} \hline
\textbf{Dataset} &\textbf{Category} & \multicolumn{4}{|l|}{\textbf{Faster R-CNN}} & \multicolumn{4}{|l|}{\textbf{Mask R-CNN}} \\ \cline{3-10} 
& &\textbf{R}$\uparrow$ &\textbf{P}$\uparrow$ &\textbf{F}$\uparrow$ &\textbf{mAP}$\uparrow$ &\textbf{R}$\uparrow$ &\textbf{P}$\uparrow$ &\textbf{F}$\uparrow$ &\textbf{mAP}$\uparrow$ \\ \hline
Validation &Table &0.9571 &0.9260 &0.9416 &0.9554 &\textbf{0.9824} &\textbf{0.9664} &\textbf{0.9744} &\textbf{0.9761} \\ \cline{2-10}
           &Figure &0.8607 &0.7800 &0.8204 &0.8103 &\textbf{0.8699} &\textbf{0.8326} &\textbf{0.8512} &\textbf{0.8391} \\ \cline{2-10}
           &Natural Image &0.9027 &0.8607 &0.8817 &0.8803 &\textbf{0.9461} &\textbf{0.8820} &\textbf{0.9141} &\textbf{0.9174}\\ \cline{2-10}
           &Logo &0.8566 &0.4063 &0.6315 &0.6217 &\textbf{0.8852} &\textbf{0.4122} &\textbf{0.6487} &\textbf{0.6434} \\ \cline{2-10}
           &Signature &0.9411 &0.8000 &0.8705 &0.9135 &\textbf{0.9633} &\textbf{0.8400} &\textbf{0.9016} &\textbf{0.9391} \\ \cline{2-10}
           &Average &0.9026 &0.7546 &0.8291 &0.8362 &\textbf{0.9294} &\textbf{0.7866} &\textbf{0.8580} &\textbf{0.8630} \\ \hline    
Test       &Table &0.9512 &0.9234 &0.9373 &0.9392 &\textbf{0.9711} &\textbf{0.9715} &\textbf{0.9713} &\textbf{0.9654} \\ \cline{2-10}
           &Figure &0.8530 &0.7582 &0.8056 &0.8332 &\textbf{0.8898} &\textbf{0.7872} &\textbf{0.8385} &\textbf{0.8686} \\ \cline{2-10}
           &Natural Image &0.8745 &0.8631 &0.8688 &0.8445 &\textbf{0.9179} &\textbf{0.8625} &\textbf{0.8902} &\textbf{0.8945} \\ \cline{2-10}
           &Logo &0.5933 &0.3606 &0.4769 &0.4330 &\textbf{0.6330} &\textbf{0.3920} &\textbf{0.5125} &\textbf{0.4699} \\ \cline{2-10}
           &Signature &0.8868 &0.7753 &0.8310 &0.8981 &\textbf{0.9175} &\textbf{0.7876} &\textbf{0.8525} &\textbf{0.9115} \\ \cline{2-10}
           &Average &0.8318 &0.7361 &0.7839 &0.7896 &\textbf{0.8659} &\textbf{0.7601} &\textbf{0.8130} &\textbf{0.8220} \\ \hline 
\end{tabular}
\end{center}
\caption{Quantitative results of two baseline approaches for detecting graphical objects. \textbf{R:} indicates Recall. \textbf{P:} indicates Precision. \textbf{F:} indicates {\sc f}-measure. \textbf{mAP:} indicates mean average precision. \label{table_quantitative_result}} 
\end{table}

\subsection{Effectiveness of IIIT-AR-13K Over Existing Larger Datasets}

The newly generated dataset {\sc iiit-ar-13k} is smaller (with respect to number of document images) than some of the existing datasets (e.g., {\sc d}eep{\sc f}igures, {\sc t}able{\sc b}ank, and {\sc p}ub{\sc l}ay{\sc n}et) for the graphical object (i.e., table) detection task. We establish the effectiveness of the smaller {\sc iiit-ar-13k} dataset over the larger existing datasets - {\sc d}eep{\sc f}igures, {\sc t}able{\sc b}ank, and {\sc p}ub{\sc l}ay{\sc n}et for graphical object (i.e., table common object category of all datasets) detection task. In~\cite{tablebank}, the authors experimentally show that {\sc t}able{\sc b}ank dataset is more effective than the largest {\sc d}eep{\sc f}igures dataset for table detection. In this paper, we use {\sc t}able{\sc b}ank and {\sc p}ub{\sc l}ay{\sc n}et datasets to establish the effectiveness of the newly generated dataset over the existing datasets for table detection. For this purpose, {\sc m}ask {\sc r-cnn} model is trained with training set of each of three datasets - {\sc t}able{\sc b}ank, {\sc p}ub{\sc l}ay{\sc n}et, and {\sc iiit-ar-13k}. These trained models are individually evaluated on test set of existing datasets - {\sc icdar-2013}, {\sc icdar-pod-2017}, c{\sc td}a{\sc r}, {\sc unlv}, {\sc m}armot, {\sc p}ub{\sc l}ay{\sc n}et. We name these experiments as 

\begin{enumerate}[(i)]
\item \textbf{Experiment-I:} {\sc m}ask {\sc r-cnn} trained with the {\sc t}able{\sc b}ank ({\sc l}a{\sc t}ex) dataset.
\item \textbf{Experiment-II:} {\sc m}ask {\sc r-cnn} trained with the {\sc t}able{\sc b}ank ({\sc w}ord) dataset.
\item \textbf{Experiment-III:} {\sc m}ask {\sc r-cnn} trained with the {\sc t}able{\sc b}ank ({\sc l}a{\sc t}ex+{\sc w}ord) dataset. 
\item \textbf{Experiment-IV:} {\sc m}ask {\sc r-cnn} trained with the {\sc p}ub{\sc l}ay{\sc n}et dataset.
\item \textbf{Experiment-V:} {\sc m}ask {\sc r-cnn} trained with the {\sc iiit-ar-13k} dataset. 
\end{enumerate}

\begin{table}[ht!]
\begin{center}
\begin{tabular}{|l|l|l|l|l |l|} \hline
\textbf{Test Dataset} &\textbf{Training Dataset} &\multicolumn{4}{|l|}{\textbf{Quantitative Score}} \\ \cline{3-6} 
& & \textbf{R}$\uparrow$ &\textbf{P}$\uparrow$ &\textbf{F}$\uparrow$ &\textbf{mAP}$\uparrow$ \\ \hline
{\sc icdar-2013} &{\sc tbl} &0.9454 &0.9341 &\textbf{0.9397} &0.9264 \\
                 &{\sc tbw} &0.9090 &\textbf{0.9433} &0.9262 &0.8915 \\
                 &{\sc tblw}&0.9333 &0.9277 &0.9305 &0.9174 \\
                 &{\sc p}ub{\sc l}ay{\sc n}et &0.9272 &0.8742 &0.9007 &0.9095 \\
                 &{\sc iiit-ar-13k}&\textbf{0.9575} &0.8977 &0.9276 &\textbf{0.9393} \\ \hline

{\sc icdar-2017} &{\sc tbl}  &0.9274 &0.7016 &0.8145 &0.8969 \\
                &{\sc tbw}   &0.8107 &0.5908 &0.7007 &0.7520 \\
                &{\sc tblw}  &\textbf{0.9369} &\textbf{0.7406} &\textbf{0.8387} &\textbf{0.9035} \\
                &{\sc p}ub{\sc l}ay{\sc n}et &0.8296 &0.6368 &0.7332 &0.7930 \\
                 &{\sc iiit-ar-13k} &0.8675 &0.6311 &0.7493 &0.7509 \\ \hline 
                 
c{\sc td}a{\sc r} &{\sc tbl}  &0.7006 &0.7208 &0.71078 &0.5486 \\
                 &{\sc tbw}   &0.6405 &0.7582 &0.6994 &0.5628 \\
                 &{\sc tblw}  &0.7230 &0.7481 &0.7356 &0.5922 \\
                 &{\sc p}ub{\sc l}ay{\sc n}et &0.6391 &0.7231 &0.6811 &0.5610 \\
                 &{\sc iiit-ar-13k}&\textbf{0.8097} &\textbf{0.8224} &\textbf{0.8161} &\textbf{0.7478} \\ \hline
                 
{\sc unlv}       &{\sc tbl}  &0.5806 &0.6612 &0.6209 &0.5002 \\
                 &{\sc tbw}  &0.5035 &0.8005 &0.6520 &0.4491 \\
                 &{\sc tblw} &0.6362 &0.6787 &0.6574 &0.5513 \\
                 &{\sc p}ub{\sc l}ay{\sc n}et  &0.7329 &0.8329 &0.7829 &0.6950 \\
                 &{\sc iiit-ar-13k} &\textbf{0.8602} &\textbf{0.7843} &\textbf{0.8222} &\textbf{0.7996} \\ \hline          
                 
{\sc m}armot  &{\sc tbl}   &0.8860 &\textbf{0.8840} &0.8850 &\textbf{0.8465}  \\      
                 &{\sc tbw}   &0.8423 &0.8808 &0.8615 &0.8026 \\
                 &{\sc tblw}  &\textbf{0.8919} &0.8802 &\textbf{0.8860} &0.8403 \\
                 &{\sc p}ub{\sc l}ay{\sc n}et  &0.8549 &0.8214 &0.8382 &0.7987 \\
                 &{\sc iiit-ar-13k} &0.8852 &0.8075 &0.8464 &0.8464 \\ \hline
\end{tabular}
\end{center}
\caption{Performance of table detection in the existing datasets. Model is trained with only training images containing tables of the respective datasets. \textbf{TBL:} indicates {\sc t}able{\sc b}ank ({\sc l}a{\sc t}ex). \textbf{TBW:} indicates {\sc t}able{\sc b}ank ({\sc w}ord). \textbf{TBLW:} indicates {\sc t}able{\sc b}ank ({\sc l}a{\sc t}ex+{\sc w}ord).  \textbf{R:} indicates Recall. \textbf{P:} indicates Precision. \textbf{F:} indicates {\sc f}-measure. \textbf{mAP:} indicates mean average precision. \label{table_effectiveness_result}} 
\end{table}
\vspace{-0.01\textwidth}

\paragraph{\textbf{Observation-I - Without Fine-tuning:}}

To establish effectiveness of the newly generated {\sc iiit-ar-13k} dataset over larger existing datasets, we do experiments: Experiment-I, Experiment-II, Experiment-III, Experiment-IV, and Experiment-V. Table~\ref{table_effectiveness_result} illustrates quantitative results of these experiments. The table highlights that Experiment-V produces the best results (with respect to {\sc f}-measure and m{\sc ap}) for c{\sc td}a{\sc r}, {\sc unlv}, and {\sc iiit-ar-13k} datasets. Though Experiment-I produces the best m{\sc ap} value for {\sc m}armot dataset, the performance of Experiment-V on this dataset is very close to the performance of Experiment-I. For {\sc icdar-2017} dataset, the performance of Experiment-V is lower than Experiment-I and Experiment-III with respect to {\sc f}-measure. The table highlights that, even though {\sc iiit-ar-13k} is significantly smaller than {\sc t}able{\sc b}ank and {\sc p}ub{\sc l}ay{\sc n}et, it is more effective for training a model for detecting tables in the existing datasets.

\begin{figure}[ht!]
\centerline{
\tcbox[sharp corners, size = tight, boxrule=0.4mm, colframe=black, colback=white]{
\epsfig{figure=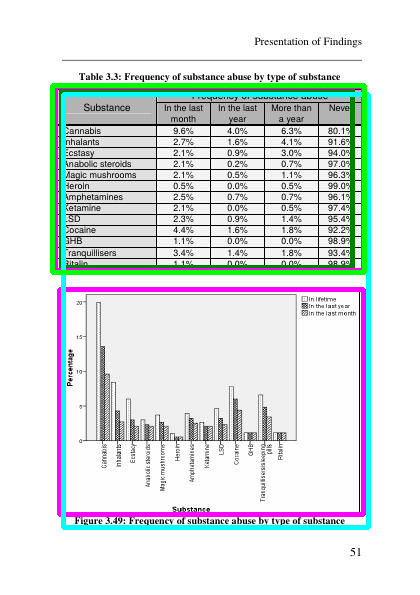, width=0.25\textwidth,height=0.16\textwidth}}
\hspace{-0.01\textwidth}
\tcbox[sharp corners, size = tight, boxrule=0.4mm, colframe=black, colback=white]{
\epsfig{figure=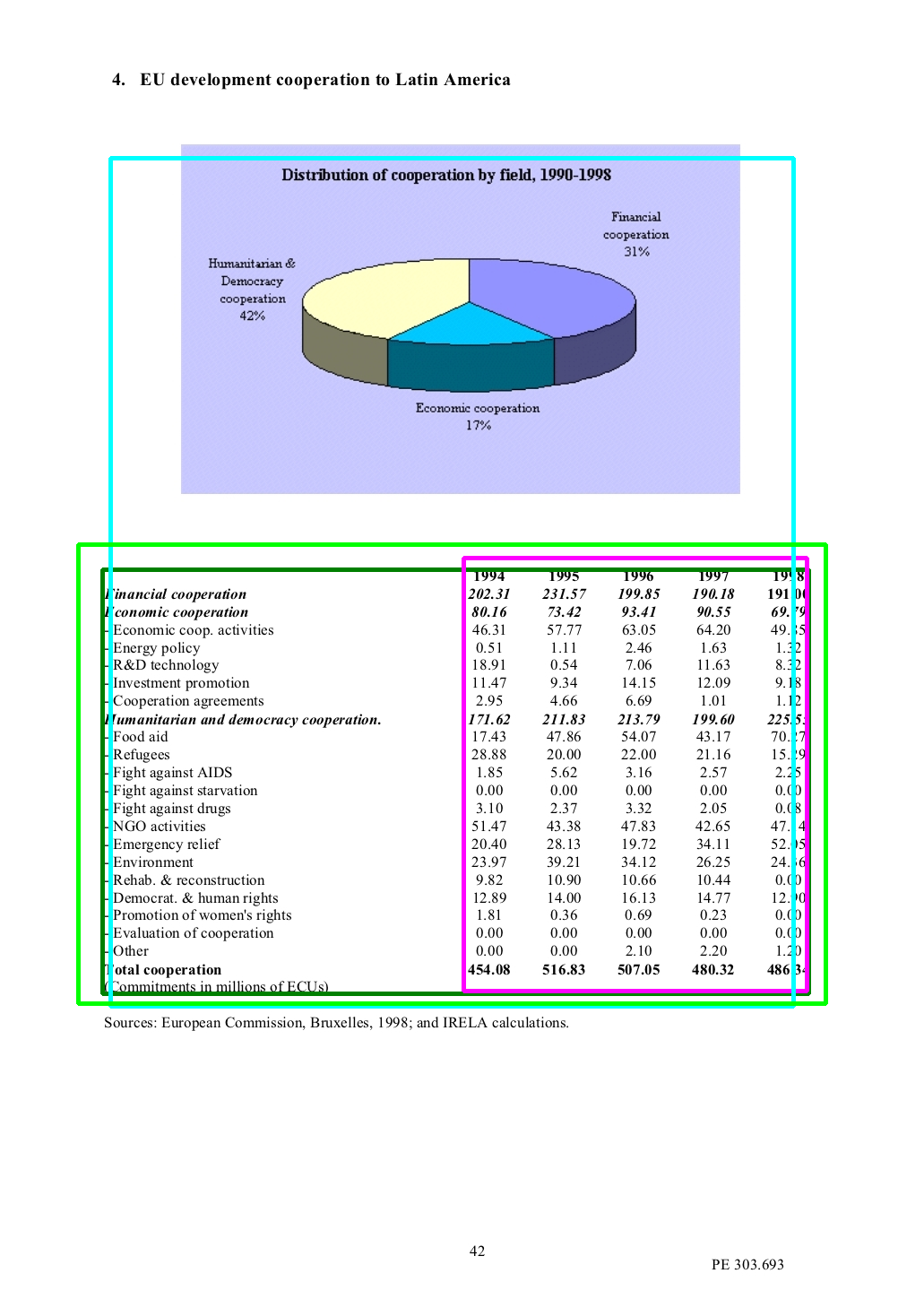, width=0.25\textwidth,height=0.16\textwidth}}
\hspace{-0.01\textwidth}
\tcbox[sharp corners, size = tight, boxrule=0.4mm, colframe=black, colback=white]{
\epsfig{figure=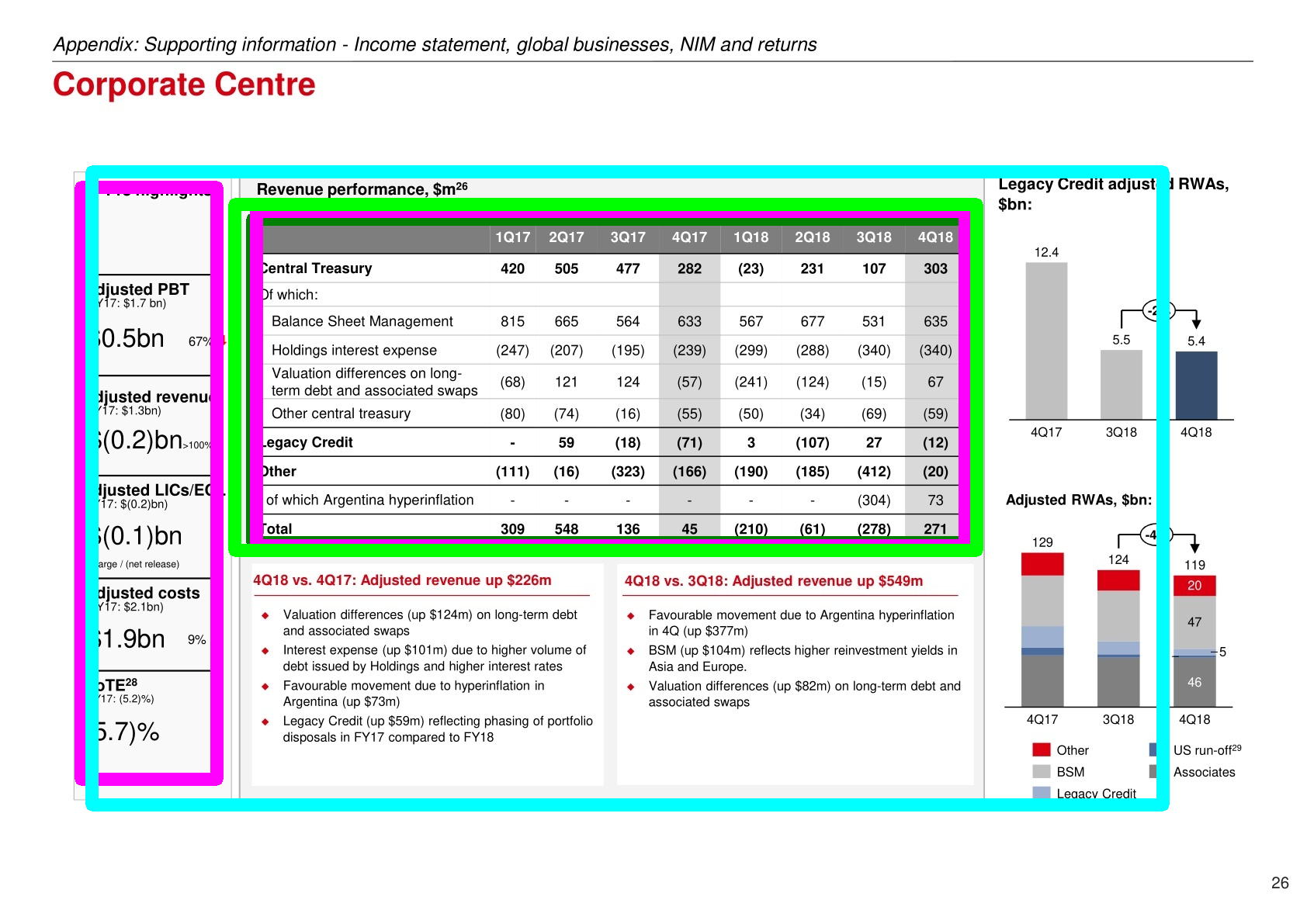, width=0.25\textwidth,height=0.16\textwidth}}
\hspace{-0.01\textwidth}
\tcbox[sharp corners, size = tight, boxrule=0.4mm, colframe=black, colback=white]{
\epsfig{figure=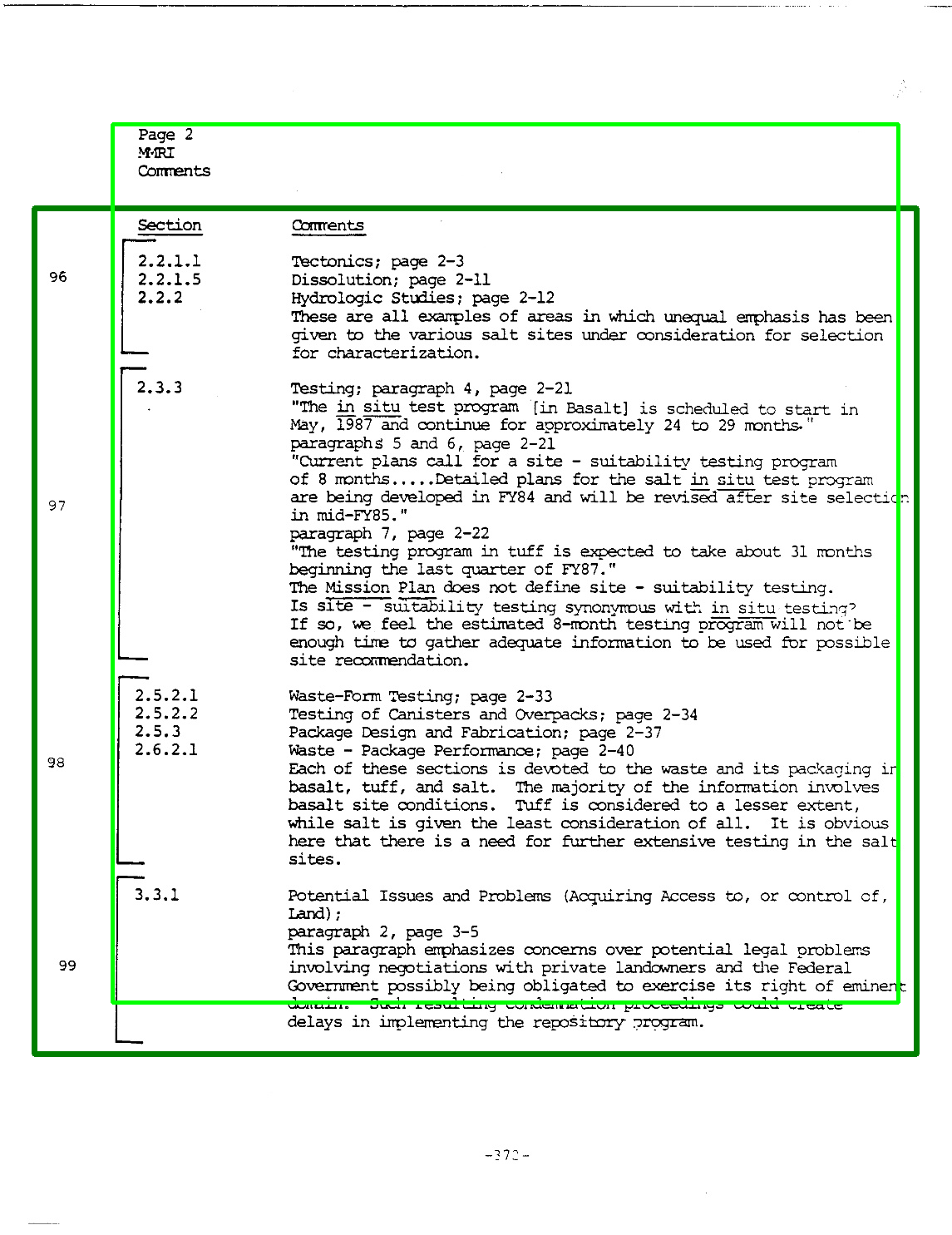, width=0.25\textwidth,height=0.16\textwidth}}}
\vspace{-0.03\textwidth}
\centerline{(a) \hspace{0.2\textwidth} (b) \hspace{0.2\textwidth} (c) \hspace{0.2\textwidth} (d)}
\vspace{0.01\textwidth}
\centerline{
\tcbox[sharp corners, size = tight, boxrule=0.4mm, colframe=black, colback=white]{
\epsfig{figure=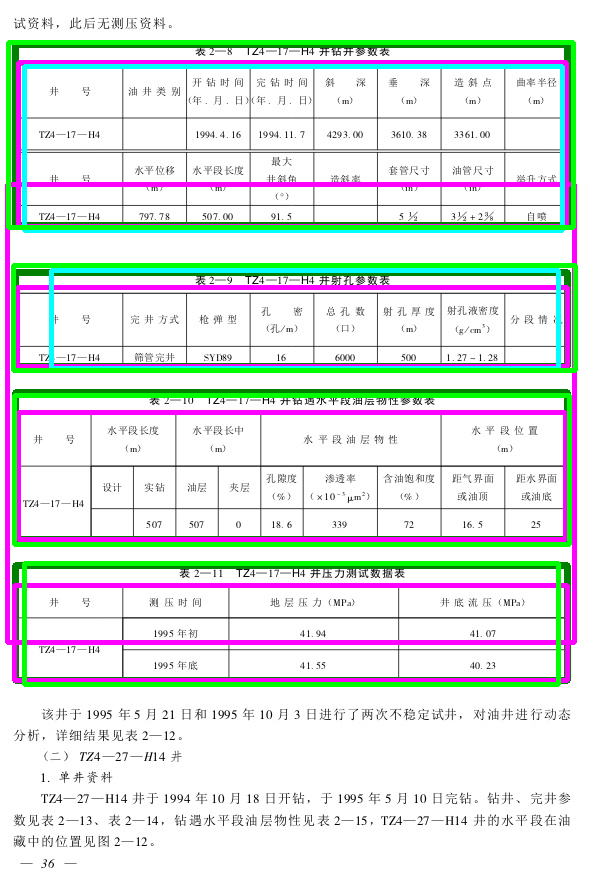, width=0.25\textwidth,height=0.16\textwidth}}
\hspace{-0.01\textwidth}
\tcbox[sharp corners, size = tight, boxrule=0.4mm, colframe=black, colback=white]{
\epsfig{figure=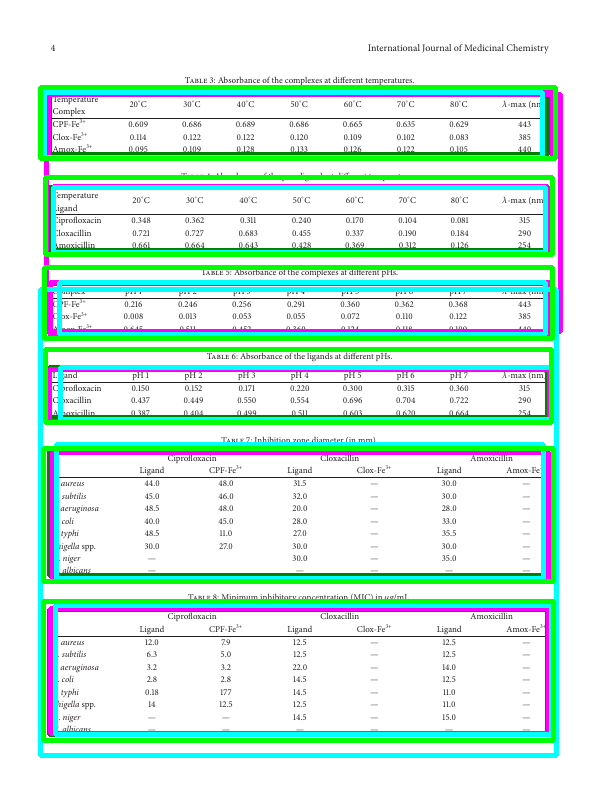, width=0.25\textwidth,height=0.16\textwidth}}
\hspace{-0.01\textwidth}
\tcbox[sharp corners, size = tight, boxrule=0.4mm, colframe=black, colback=white]{
\epsfig{figure=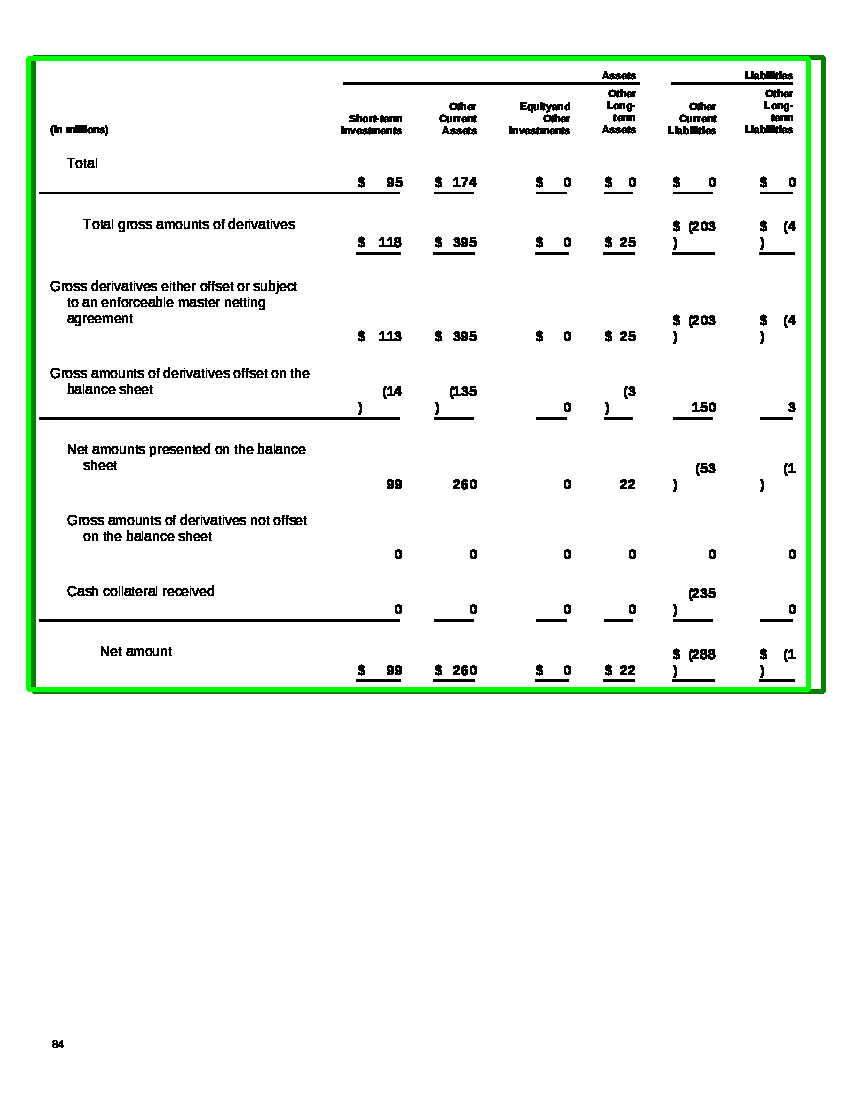, width=0.25\textwidth,height=0.16\textwidth}}
\hspace{-0.01\textwidth}
\tcbox[sharp corners, size = tight, boxrule=0.4mm, colframe=black, colback=white]{
\epsfig{figure=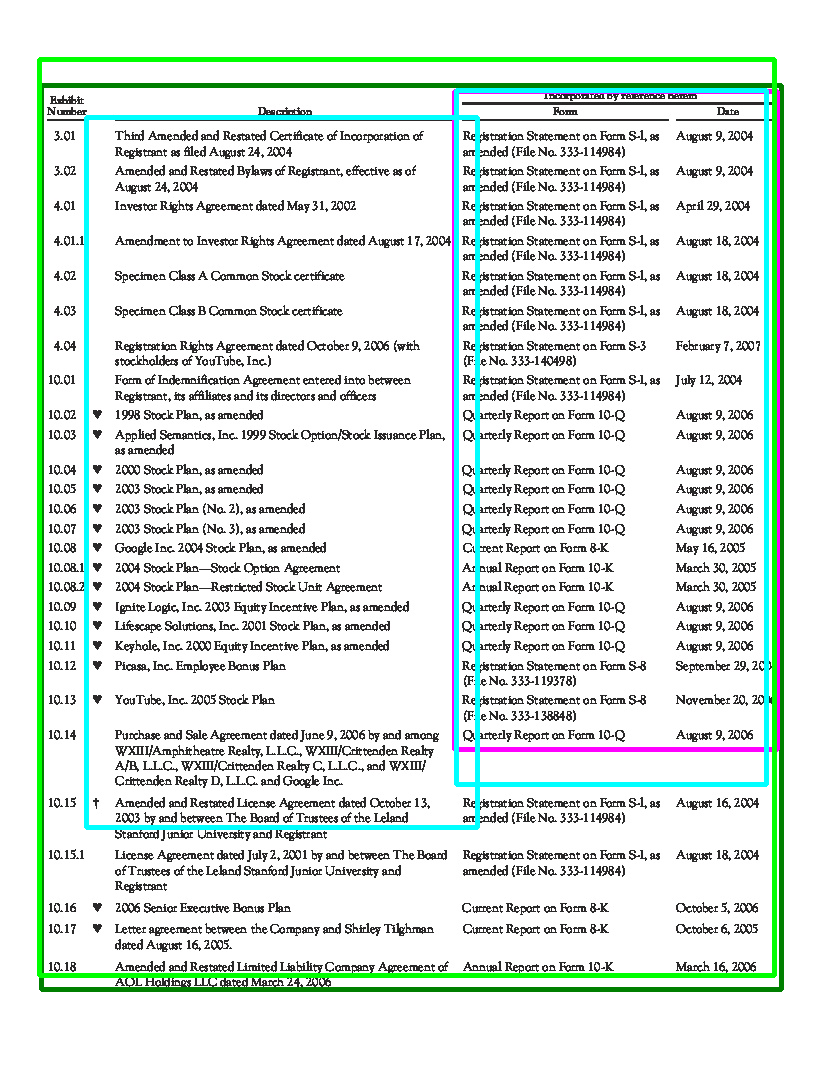, width=0.25\textwidth,height=0.16\textwidth}}}
\vspace{-0.03\textwidth}
\centerline{(e) \hspace{0.2\textwidth} (f) \hspace{0.2\textwidth} (g) \hspace{0.2\textwidth} (h)}
\caption{Few examples of predicted bounding boxes of tables in various datasets - (a) {\sc icdar-2013}, (b) {\sc icadr-pod-2017}, (c) c{\sc td}a{\sc r}, (d) {\sc unlv}, (e) {\sc m}armot, (f) {\sc p}ub{\sc l}ay{\sc n}et, (g) {\sc iiit-ar-13k} (validation), (h) {\sc iiit-ar-13k} (test) using \textbf{Experiment-I}, \textbf{Experiment-IV} and \textbf{Experiment-V}. \textbf{Experiment-I:} {\sc m}ask {\sc r-cnn} trained with {\sc t}able{\sc b}ank ({\sc l}a{\sc t}ex) dataset. \textbf{Experiment-IV:} {\sc m}ask {\sc r-cnn} trained with {\sc p}ub{\sc l}ay{\sc n}et dataset. \textbf{Experiment-V:} {\sc m}ask {\sc r-cnn} trained with {\sc iiit-ar-13k} dataset. \textbf{Dark Green:} rectangle highlights the ground truth bounding boxes of tables. \textbf{Pink, Cyan, and Light Green:} rectangles indicate the predicted bounding boxes of tables using \textbf{Experiment-I}, \textbf{Experiment-IV} and \textbf{Experiment-V}, respectively. \label{fig_failure}}
\end{figure}

Figure~\ref{fig_failure} shows the predicted bounding boxes of tables in the pages of several datasets using three different models. Here, dark Green, Pink, Cyan, and Light Green colored rectangles highlighted the ground truth and predicted bounding boxes of tables using Experiment-I, Experiment-IV, and Experiment-V, respectively. In those images, Experiment-V correctly detects the tables. At the same time, either Experiment-I or Experiment-IV or Experiment-I and Experiment-IV fails to predict the bounding boxes of the tables accurately.

\paragraph{\textbf{Observation-II - Fine-tuning with Complete Training Set:}}

From Table~\ref{table_effectiveness_result2}, it is also observed that the performance of Experiment-I, Experiment-II, Experiment-III, and Experiment-IV on the validation set and test set of {\sc iiit-ar-13k} dataset is significantly worse (10\% in case of {\sc f}-measure and 15\% in case of m{\sc ap}) than the performance of Experiment-V. On the contrary, when we fine-tune with the training set of {\sc iiit-ar-13k} dataset, all the fine-tuning experiments obtain similar outputs compared to Experiment-V. In the case of the validation set of {\sc p}ub{\sc l}ay{\sc n}et dataset, Experiment-IV outperforms Experiment-II and Experiment-V; and the performances of Experiment-I and Experiment-III are very close to the performance of Experiment-IV. When we fine-tune with the training set of {\sc p}ub{\sc l}ay{\sc n}et dataset, the fine-tuned model achieves similar performance compared to Experiment-IV. These experiments highlight that fine-tuning with the complete training set is effective for both the larger existing datasets and as well as {\sc iiit-ar-13k} dataset.

\begin{table}
\begin{center}
\addtolength{\tabcolsep}{-1.5pt}
\begin{tabular}{|l|l|l| l| l| l |l|} \hline
\textbf{Test Dataset} &\textbf{Training Dataset} &\textbf{Fine-tune} &\multicolumn{4}{|l|}{\textbf{Quantitative Score}} \\ \cline{4-7} 
& & &\textbf{R}$\uparrow$ &\textbf{P}$\uparrow$ &\textbf{F}$\uparrow$ &\textbf{mAP}$\uparrow$ \\ \hline

{\sc p}ub{\sc l}ay{\sc n}et  &{\sc tbl} & &0.9859 &0.6988 &0.8424 &0.9461\\
(validation)    &{\sc tbw}  & &0.9056 &0.7194 &0.8125 &0.8276 \\
                &{\sc tblw} & &0.9863 &0.7258 &0.8560 &0.9557 \\
                &{\sc p}ub{\sc l}ay{\sc n}et & &\textbf{0.9886} &0.7780 &0.8833 &0.9776 \\
                &{\sc iiit-ar-13k} & &0.9555 &0.5530 &0.7542 &0.8696  \\ \cline{2-7}
                &{\sc iiit-ar-13k} &{\sc p}ub{\sc l}ay{\sc n}et({\sc f-trs}) &0.9882 &\textbf{0.7898} &\textbf{0.8890} &\textbf{0.9781}   \\ \cline{2-7}
                &{\sc iiit-ar-13k} &{\sc p}ub{\sc l}ay{\sc n}et({\sc p-trs}) &0.9869 &0.4653 &0.7261 &0.9712   \\ \hline
                 
{\sc iiit-ar-13k} &{\sc tbl} &  &0.8109 &0.7370 &0.7739 &0.7533 \\
 (validation)    &{\sc tbw} & &0.7641 &0.8214 &0.7928 &0.7230 \\
                 &{\sc tblw} & &0.8217 &0.7345 &0.7781 &0.7453 \\ 
                 &{\sc p}ub{\sc l}ay{\sc n}et & &0.8100 &0.7092 &0.7596 &0.7382 \\
                 &{\sc iiit-ar-13k} &  &\textbf{0.9891} &0.7998 &0.8945 &0.9764 \\ \cline{2-7}
                 
          &{\sc tbl}  &{\sc iiit-ar-13k}({\sc f-trs}) &0.9869 &0.8600 &0.9234 &0.9766 \\
          &{\sc tbw}  & &0.9815 &0.8065 &0.8940 &0.9639 \\
                 &{\sc tblw}      & &0.9873 &0.8614 &0.9244 &\textbf{0.9785} \\ 
                 &{\sc p}ub{\sc l}ay{\sc n}et  & &0.9842 &0.8344 &0.9093 &0.9734 \\\cline{2-7}
     &{\sc tbl}  &{\sc iiit-ar-13k}({\sc p-trs}) &0.9747 &0.8790 &0.9269 &0.9648 \\
     &{\sc tbw}    & &0.9734 &0.8767 &0.9251 &0.9625 \\
     &{\sc tblw}      & &0.9783 &\textbf{0.8998} &\textbf{0.9391} &0.9717 \\ 
     &{\sc p}ub{\sc l}ay{\sc n}et  & &0.9797 &0.8732 &0.9264 &0.9687 \\\hline     
                 
 {\sc iiit-ar-13k} &{\sc tbl} & &0.8023 &0.7428 &0.7726 &0.7400 \\       
 (test)          &{\sc tbw} & &0.7704 &0.8396 &0.8050 &0.7276 \\
                 &{\sc tblw} & &0.8278 &0.7500 &0.7889 &0.7607 \\
                 &{\sc p}ub{\sc l}ay{\sc n}et & &0.8093 &0.7302 &0.7697 &0.7313 \\
                 &{\sc iiit-ar-13k} & &\textbf{0.9826} &0.8361 &0.9093 &0.9688 \\ \cline{2-7}
 &{\sc tbl} &{\sc iiit-ar-13k}({\sc f-trs}) &0.9761 &0.8774 &0.9267 &0.9659 \\       
 &{\sc tbw}  & &0.9753 &0.8239 &0.8996 &0.9596 \\
 &{\sc tblw} & &0.9776 &0.8788 &0.9282 &\textbf{0.9694} \\
 &{\sc p}ub{\sc l}ay{\sc n}et & &0.9753 &0.8571 &0.9162 &0.9642  \\ \cline{2-7}
 &{\sc tbl} &{\sc iiit-ar-13k}({\sc p-trs}) &0.9637 &0.9022 &0.9330 &0.9525 \\       
 &{\sc tbw}  & &0.9699 &0.8922 &0.9311 &0.9615 \\
 &{\sc tblw} & &0.9718 &\textbf{0.9091} &\textbf{0.9405} &0.9629 \\
 &{\sc p}ub{\sc l}ay{\sc n}et & &0.9684 &0.8905 &0.9294 &0.9552  \\ \hline   
\end{tabular}
\end{center}
\caption{Performance of table detection using fine-tuning. \textbf{TBL:} indicates {\sc t}able{\sc b}ank ({\sc l}a{\sc t}ex). \textbf{TBW:} indicates {\sc t}able{\sc b}ank ({\sc w}ord). \textbf{TBLW:} indicates {\sc t}able{\sc b}ank ({\sc l}a{\sc t}ex$+${\sc w}ord). \textbf{F-TRS:} indicates complete training set. \textbf{P-TRS:} indicates partial training set i.e., only 1{\sc k} randomly selected training images. \textbf{R:} indicates Recall. \textbf{P:} indicates Precision. \textbf{F:} indicates {\sc f}-measure. \textbf{mAP:} indicates mean average precision. \label{table_effectiveness_result2}} 
\end{table}

\vspace{-0.01\textwidth}
\paragraph{\textbf{Observation-III - Fine-tuning with Partial Training Set:}}

From Table~\ref{table_effectiveness_result2}, we observe that the performance of the trained model using {\sc t}able{\sc b}ank and {\sc p}ub{\sc l}ay{\sc n}et on validation and test sets of {\sc iiit-ar-13k} dataset is (10-20\%) less than the performance of model trained with training set of {\sc iiit-ar-13k}. When we fine-tune the models with a partial training set of only 1{\sc k} randomly selected images of {\sc iiit-ar-13k}, the models also achieve similar outputs (see Table~\ref{table_effectiveness_result2}). 
On the other hand, the performance of the trained model with {\sc iiit-ar-13k} dataset on validation set of {\sc p}ub{\sc l}ay{\sc n}et is (10-15\%) is less than models trained with {\sc t}able{\sc b}ank and {\sc p}ub{\sc l}ay{\sc n}et. When we fine-tune the model (trained with {\sc iiit-ar-13k} dataset) with the complete training set of {\sc p}ub{\sc l}ay{\sc n}et, the model obtains very close output to the models trained with {\sc t}able{\sc b}ank and {\sc p}ub{\sc l}ay{\sc n}et. But when we fine-tune the model (trained with {\sc iiit-ar-13k} dataset) with the partial training set (only 1{\sc k} images) of {\sc p}ub{\sc l}ay{\sc n}et, the model is unable to obtain similar output (with respect to {\sc f}-measure) to the models trained with {\sc t}able{\sc b}ank and {\sc p}ub{\sc l}ay{\sc n}et. These experiments also highlight that the newly created {\sc iiit-ar-13k} dataset is more effective for fine-tuning.

\vspace{-0.02\textwidth}
\section{Summary and Observations}

This paper presents a new dataset {\sc iiit-ar-13k} for detecting graphical objects in business documents, specifically annual reports. It consists of 13{\sc k} pages of annual reports of more than ten years of twenty-nine different companies with bounding box annotation of five different object categories --- table, figure, natural image, logo, and signature. 
\begin{itemize}
\item The newly generated dataset is the largest manually annotated dataset for graphical object detection purpose, at this stage. 
\item This dataset has more labels and diversity compared to most of the existing datasets.
\item Though the {\sc iiit-ar-13k} is smaller than the existing automatic annotated datasets --- {\sc d}eep{\sc f}igures, {\sc p}ub{\sc l}ay{\sc n}et, and {\sc t}able{\sc n}et, the model trained with {\sc iiit-ar-13k} performs better than the model trained with the larger datasets for detecting tables in document images in most cases. 
\item Models trained with the existing datasets also achieve better performance by fine-tuning with a limited number of training images from {\sc iiit-ar-13k}. 
\end{itemize}

We believe that this dataset will aid research in detecting tables and other graphical objects in business documents. 
\bibliographystyle{splncs04}
\bibliography{reference}
\end{document}